\documentclass[10pt,twocolumn,letterpaper]{article}

\usepackage{iccv}

\usepackage{times,latexsym}
\usepackage{microtype}
\usepackage{graphicx} \graphicspath{{figures/}}
\usepackage{amsmath,amssymb,mathabx,amsbsy,mathtools,etoolbox,mathbbol}
\usepackage[linesnumbered,ruled,vlined]{algorithm2e}
\usepackage{acronym}
\usepackage{enumitem}
\usepackage{balance}
\usepackage{xspace}
\usepackage{setspace}
\usepackage[skip=3pt,font=small]{subcaption}
\usepackage[skip=3pt,font=small]{caption}
\usepackage{booktabs,tabularx,colortbl,multirow,array,makecell}
\usepackage{overpic,wrapfig}
\usepackage{dblfloatfix}
\usepackage{nicefrac}
\usepackage[misc]{ifsym}
\usepackage{siunitx} 
\usepackage{soul} 
\usepackage[pagebackref,breaklinks,colorlinks,bookmarks=false]{hyperref}
\usepackage[capitalise,nameinlink]{cleveref}

\makeatletter
\DeclareRobustCommand\onedot{\futurelet\@let@token\@onedot}
\def\@onedot{\ifx\@let@token.\else.\null\fi\xspace}
\def\eg{\emph{e.g}\onedot} 

\def\ie{\emph{i.e}\onedot}

\def\vs{\emph{vs}\onedot}

\def\etal{\emph{et al}\onedot}
\makeatother

\makeatletter
\def\BState{\State\hskip-\ALG@thistlm}
\makeatother

\makeatletter
\renewcommand{\paragraph}{%
  \@startsection{paragraph}{4}%
  {\z@}{0ex \@plus 0ex \@minus 0ex}{-1em}%
  {\hskip\parindent\normalfont\normalsize\bfseries}%
}
\makeatother

\crefname{algorithm}{Alg.}{Algs.}
\Crefname{algocf}{Algorithm}{Algorithms}
\crefname{section}{Sec.}{Secs.}
\Crefname{section}{Section}{Sections}
\crefname{table}{Tab.}{Tabs.}
\Crefname{table}{Table}{Tables}
\crefname{figure}{Fig.}{Fig.}
\Crefname{figure}{Figure}{Figure}

\acrodef{lai}[LAI]{label affinity inference}
\acrodef{pce}[PCE]{partial binary cross-entropy}
\acrodef{crf}[CRF]{conditional random field}
\acrodef{em}[EM]{expectation--maximization}
\acrodef{metric}[mIoU]{mean intersection-over-union}

\ificcvfinal\fi
\ificcvfinal\fi

\iccvfinalcopy 

\begin{document}

\title{STRAP: \underline{S}tructured \underline{Tr}ansformer for\\\underline{A}ffordance Segmentation with \underline{P}oint Supervision%
}

\author{
Leiyao Cui\\
Beijing Institute of Technology\\
{\tt\small cuileiyaony@gmail.com}
\and
Xiaoxue Chen, Hao Zhao, Guyue Zhou\\
AIR, Tsinghua University\\
{\tt\small zhaohao@air.tsinghua.edu.cn}
\and
Yixin Zhu\\
Peking University\\
{\tt\small yixin.zhu@ucla.edu}
}

\maketitle

\begin{abstract}
With significant annotation savings, point supervision has been proven effective for numerous 2D and 3D scene understanding problems. This success is primarily attributed to the structured output space; \ie, samples with high spatial affinity tend to share the same labels. Sharing this spirit, we study affordance segmentation with point supervision, wherein the setting inherits an unexplored dual affinity---spatial affinity and label affinity. By \textbf{label affinity}, we refer to affordance segmentation as a multi-label prediction problem: A plate can be both \textit{holdable} and \textit{containable}. By \textbf{spatial affinity}, we refer to a universal prior that nearby pixels with similar visual features should share the same point annotation. To tackle label affinity, we devise a dense prediction network that enhances label relations by effectively densifying labels in a new domain (\ie, label co-occurrence). To address spatial affinity, we exploit a Transformer backbone for global patch interaction and a regularization loss. In experiments, we benchmark our method on the challenging CAD120 dataset, showing significant performance gains over prior methods. We provide access to the codebase of our work at \url{https://github.com/LeiyaoCui/STRAP}.
\end{abstract}
\section{Introduction}

Computer vision has extensively explored the synergy between low- and high-level affinity. For example, superpixels---the perceptual grouping of pixels---can facilitate object segmentation with much fewer computational efforts: Two nearby pixels likely belong to the same object (high-level affinity) if they carry similar colors (low-level affinity). Sharing a similar spirit, Graph Cuts~\cite{boykov2001fast}, a representative method in segmentation, propagates the source's label and a sink point to the entire image for globally optimal segmentation. In the era of deep learning, this prior has been rejuvenated by pointly-supervised methods; numerous methods~\cite{bearman2016s,cheng2022pointly,li2022fully,mcever2020pcams,tian2022vibus} explore the point-level sparse annotations to alleviate the high cost of dense annotation, showing consistent performance improvement in various settings.

\begin{figure}[ht!]
    \centering
    \includegraphics[width=\linewidth]{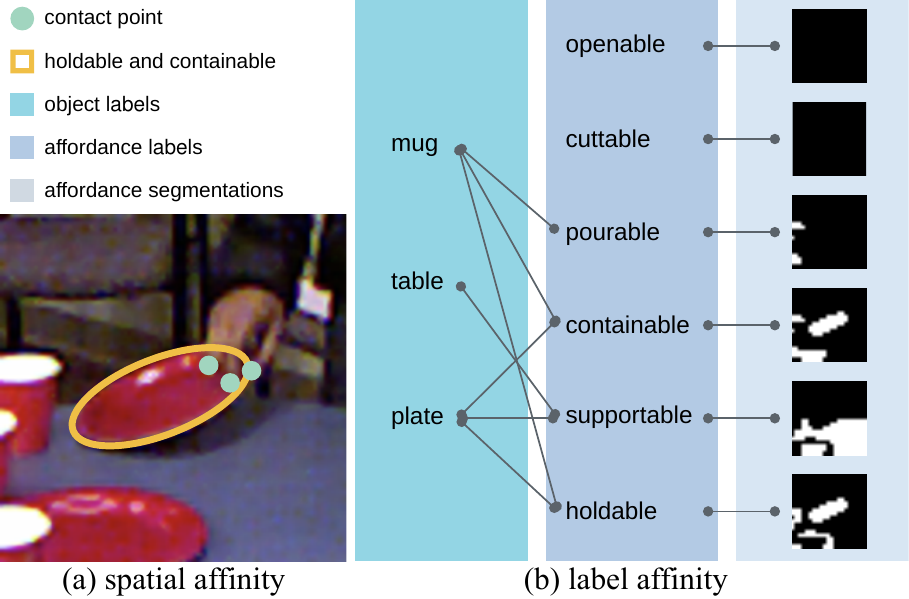}
    \caption{\textbf{Affordance segmentation with sparse point supervision.} (a) Object affordance is usually accessible by human-object interaction with sparse contact points. Propagating such sparse point annotation leverages \textbf{spatial affinity}. (b) Affordance is inherently multi-label~\cite{gibson1979ecological}: A plate can be both \textit{holdable} and \textit{containable}, thus possessing \textbf{label affinity}. By leveraging this dual affinity connecting low- and high-level image cues, we address affordance segmentation as a multi-label dense prediction problem.}
    \label{fig:teaser}
\end{figure}

In this work, we demonstrate how point supervision promotes affordance segmentation by leveraging a dual affinity that connects the low- and high-level cues in images:
\begin{itemize}[leftmargin=*,noitemsep,nolistsep]
    \item Dense affordance labels can arise from sparse point annotation by \textbf{spatial affinity}. Take \cref{fig:teaser} for an example. We can assign dense labels of \textit{holdable} to the entire plate by observing a person holding it with sparse contact points. Computationally, we can obtain sparse point annotations for the \emph{holdable} label by predicting the \emph{hold} label (action recognition) and determining the contact points between the hand and the plate (3D hand-object interaction, such as prior arts~\cite{chen2019holistic,qiu2020human}). Since the plate has a uniform color, we can propagate the sparse \emph{holdable} label from those points to the entire plate.
    \item Unlike conventional image segmentation, affordance is inherently multi-label~\cite{gibson1979ecological}; for instance, a plate is both \textit{holdable} and \textit{containable}. This unique nature of affordance touches uncharted territory in pointly-supervised segmentation: \textbf{label affinity}.
\end{itemize}

Leveraging the above unique properties introduced by dual affinity, we devise an end-to-end architecture for affordance segmentation, named \underline{S}tructured \underline{TR}ansformer for \underline{A}ffordance Segmentation with \underline{P}oint Supervision, or \textbf{STRAP} for short. To tackle label affinity, we attach a recurrent network that models pixel-wise label relations onto conventional dense prediction networks; this design allows us to densify sparse affordance labels in label space. Our recipe for spatial affinity includes two essential ingredients. The first is a Transformer backbone that models global patch interaction with self-attention, such that the long-range affinity among patches is implicitly learned. The second is a regularization loss that encourages spatially coherent predictions aligned with low-level cues.

This work makes three contributions:
\begin{itemize}[leftmargin=*,noitemsep,nolistsep]
    \item To our best knowledge, ours is the first that proposes the concept of dual affinity for affordance segmentation. Our experimental results demonstrate the efficacy of this idea.
    \item To our best knowledge, ours is the first to model label affinity for learning multi-label affordance segmentation with point-level sparse annotation.
    \item Our modeling of spatial affinity using a Transformer backbone and a regularization loss results in a compact end-to-end architecture structured in two dimensions.
\end{itemize}
\section{Related work}

\paragraph{Affordance}

The concept of affordance~\cite{gibson1979ecological} refers to an object's capability to support a specific action, which also relates to the concept of functionality~\cite{yu2011make,zhao2013scene,qi2018human,jiang2018configurable,lai2021functional}; we refer the readers to a recent survey~\cite{hassanin2021visual}. Research on affordance can be roughly categorized into three focuses. The first is \textbf{scene affordance}; the core idea is to fit the human skeleton or activities to the observed scenes~\cite{fouhey2012people,gupta20113d}, either by geometry~\cite{huang2018holistic,li2019putting} or contact force~\cite{zhu2016inferring}. Modern treatments leverage large-scale datasets by sitcoms~\cite{wang2017binge} or self-collected egocentric videos~\cite{nagarajan2020ego}. Ultimately, the goal of scene affordance is to provide valuable cues for holistic scene understanding tasks~\cite{jiang2015hallucinated,jia2020lemma,chen2021yourefit,li2023understanding,jia2022egotaskqa}. The second is \textbf{object affordance detection}~\cite{zhu2014reasoning}. With benchmark~\cite{deng20213d}, prior arts~\cite{myers2015affordance,roy2016multi,sawatzky2017weakly,chen2022cerberus} regard affordance as a dense prediction task and resolve it under pixel-level supervision. The third is affordance emerged from \textbf{human-object interactions}~\cite{wei2013modeling,zhu2015understanding,qi2018learning,mo2022o2o,jiang2022chairs}, providing interaction cues for planning and acting~\cite{edmonds2017feeling,edmonds2019tale,xu2021deep,gadre2021act,zhang2022understanding,han2022scene}. Recently, this direction has been further extended to generative models that directly synthesize interactions from observation~\cite{brahmbhatt2019contactgrasp,jiang2021hand,liu2021synthesizing,li2023gendexgrasp,wang2022humanise,huang2023diffusion}. Compared to the literature, our work addresses affordance through interactions in a pointly-supervised fashion at the pixel level, providing a trade-off between data efficiency and model performance.

\paragraph{Pointly-supervised learning}

Since dense annotation is expensive and time-consuming~\cite{richter2016playing}, the pointly-supervised approach provides a natural and effective way for weak annotation. Point supervision for semantic segmentation was first introduced~\cite{bearman2016s} for PASCAL VOC 2012 dataset~\cite{everingham2010pascal}. Some notable work includes extreme points with point-centric Gaussian channel~\cite{maninis2018deep}, a distance metric loss leveraging semantic relations among annotated points~\cite{qian2019weakly}, a point annotation scheme for uniform sampling, and applications in panoptic segmentation~\cite{mcever2020pcams,li2022fully}. In this work, we leverage the same idea of sparse point supervision for affordance segmentation.
\definecolor{resnet_color}{cmyk}{0.5474, 0.1602, 0.1944, 0}
\definecolor{transformer_encoder_color}{cmyk}{0.4268, 0.035, 0.3254, 0}
\definecolor{refine_block_color}{cmyk}{0.0449, 0.3756, 0.1817, 0}
\definecolor{fusion_block_color}{cmyk}{0.1497, 0.1317, 0.6127, 0}

\section{Method}

\paragraph{Problem definition}

We formulate affordance segmentation as a multi-label dense prediction task. We assume image $X$ contains one or more objects, and the ground-truth labels $Y$ are given by sparse point annotations to roughly locate the affordance regions. Formally, an image $X \in \mathbb{R}^{H \times W \times 3}$ has binary ground-truth labels $Y=\{y_i \mid y_i\in \mathbb{R}^{H \times W \times 1}\}$, where $y_i$ denotes the affordance label for one certain action .

\paragraph{The dual affinity property}

With only rough localizations for affordance regions, point supervision cannot accurately delineate the boundary of an affordance region. To tackle this challenge, we exploit a unique dual affinity property in affordance segmentation to compensate for the sparsity of $Y$.
Specifically, \textbf{spatial affinity} (see \cref{sec:spatial_affinity}) is the pixel similarity in terms of visual properties and positions, potentially providing additional cues to alleviate the sparsity of $Y$.
Meanwhile, \textbf{label affinity} (see \cref{sec:label_affinity}) is the structured relation between object labels and affordance labels; proper modeling of this latent space mitigates the imbalance introduced by both the sparsity of $Y$ and the long-tailed dataset.
\textbf{Computationally}, we model the dual affinity with a recurrent network (label affinity) and a regularization loss (spatial affinity).

\begin{figure*}[!ht]
    \centering
    \includegraphics[width=\linewidth]{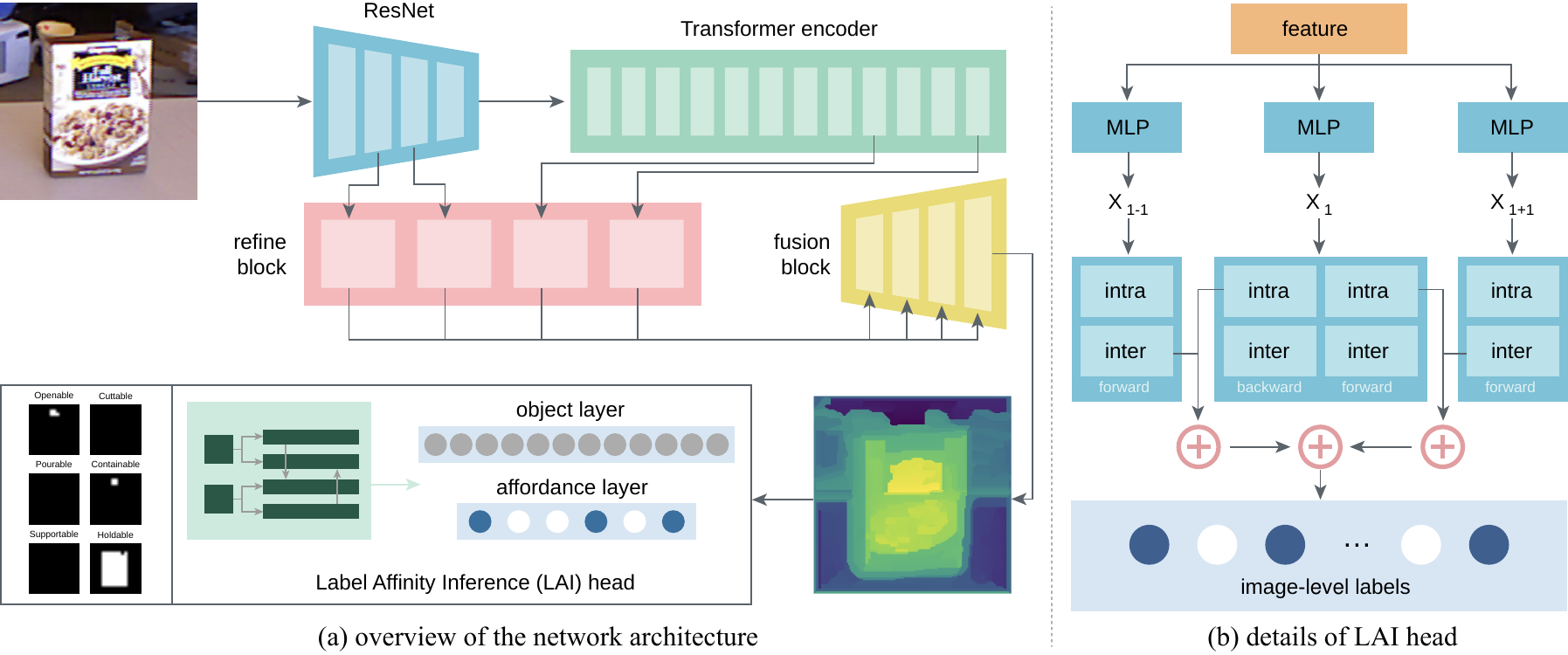}
    \caption{\textbf{Network architecture and the \acf{lai} head.}
    (a) A query image is first fed into a ResNet to extract features. Next, the feature embeddings with a position embedding are passed through Transformer encoders as tokens, reassembled from ResNet and Transformer encoders; these tokens are processed by refine blocks. Finally, a sequence of fusion blocks progressively fuses and upsamples these tokens to the image-like representations; features generated from the last fusion block are used for dense prediction and \acs{lai}.
    (b) Fully connected layers are adopted to project the given features into label affinity space, with each layer modeling the inter-relations (between affordance and object classes) and intra-relations (among different affordance classes). All label predictions in each layer are generated by a combination of top-down and bottom-up inference, and each of these inferences is a combination of inter- and intra-relations.}
    \label{fig:architecture}
\end{figure*}

\subsection{Network architecture}

The long-range dependency of Transformer-based attention mechanism is suitable for weakly-supervised methods in many applications~\cite{chen2022cerberus,gao2023semi,tian2022vibus}. In this work, we exploit a Transformer backbone with a group of dense prediction heads in a pointly-supervised manner for affordance segmentation. Specifically, we design a \acf{lai} head to model the correlations between object labels and affordance labels, pointly-supervised by human interactions with objects in daily activities. Notably, labels yielded by human interactions with objects are sparse in nature; furthermore, they differ from individuals when interacting with objects or annotating ground-truth labels. Hence, the crux of point supervision is to establish and leverage the accumulated experiences to build up the co-occurrence, therefore densifying the labels. \cref{fig:architecture}a shows network architecture.

\paragraph{Transformer backbone}

We follow the definitions and designs introduced by prior arts~\cite{chen2022cerberus,ranftl2021vision}. First, an input image $X$ is divided into $N_p = \frac{HW}{p^2}$ non-overlapping patches of size $p^2$ ($p = 16$ in our settings); We transform the list of these patches into a vector of 2-D patches. A ResNetV2-50 backbone~\cite{he2016deep} is adopted to extract feature maps as tokens (\textcolor{resnet_color}{blue $\blacksquare$} in \cref{fig:architecture}'s top left panel). Next, a learnable position embedding is concatenated to retain spatial information. Similar to the modern vision transformer designs \cite{dosovitskiy2020image,ranftl2021vision,caron2021emerging}, a special token is added to gain global information, termed the readout token. Finally, the network generates $N_p + 1$ tokens $\{t_0, t_1, \dots , t_{N_p} \}$, where $t_0$ is the readout token. All tokens are then encoded by a sequence of multi-head self-attention blocks (\textcolor{transformer_encoder_color}{green $\blacksquare$} in \cref{fig:architecture}'s top left panel). 

To reassemble the tokens from four different layers: the first and the second stage from ResNetV2-50 backbone, and the 9th and the 12th layers from Transformer encoders (\textcolor{refine_block_color}{red $\blacksquare$} in \cref{fig:architecture}'s top left panel), we design a three-stage procedure to restore the image-like representations for tokens from Transformer encoders:
\begin{enumerate}[leftmargin=*,noitemsep,nolistsep]
    \item We use a projection block to fuse the readout token $t_0$ into other tokens. The projection block is defined as
    \begin{equation}
        \small%
        \mathrm{token_{proj}}(t_i) = \mathrm{GELU}\left(\mathrm{MLP}\left(\mathrm{Concat}(t_i, t_0)\right)\right),
    \end{equation}
    where $i = 1,2,\dots,N_p$.
    \item After projection, the new $N_p$ tokens are reshaped into image-like representations from $\mathbb{R}^{N_p \times D} $ to $ \mathbb{R}^{\frac{H}{p} \times \frac{W}{p} \times D}$ by a spatial rearrangement operation.
    \item  We use a $1 \times 1$ and a $3 \times 3$ convolution to resample features from $\mathbb{R}^{\frac{H}{p} \times \frac{W}{p} \times D}$ to $\mathbb{R}^{\frac{H}{s} \times \frac{W}{s} \times \hat{D}}$, where $s>p$.
\end{enumerate}

To fuse these extracted tokens, we use RefineNet-based fusion blocks~\cite{lin2017refinenet} to upsample them progressively (\textcolor{fusion_block_color}{yellow $\blacksquare$} in \cref{fig:architecture}'s top left panel). Besides the first block, each fusion block processes the output of the previous fusion block and the related extracted token together, and the resolution of the final feature representation $F$ will be upsampled into $\mathbb{R}^{\frac{H}{2} \times \frac{W}{2} \times \hat{D}}$. The final feature representation $F$ is used for dense prediction and \ac{lai}.

\paragraph{Dense prediction head}

The dense prediction head leverages two components to generate affordance segmentation: fully connected layers to generate the affordance map and a bilinear interpolation function to restore the resolution of the affordance map to the original resolution of $X$. As such, the size of its output is upsampled into $H \times W \times 1$.

Usually, the binary cross-entropy loss is a well-behaved choice in full supervision. However, as the foreground pixel number and background pixel number are imbalanced in  point supervision, we adopt a \ac{pce} loss as the dense prediction objective, whose effectiveness is supported by the ablation studies in \cref{tab:ablation}:
\begin{equation}
    \small%
    \begin{aligned}%
        \mathcal{L}_{\mathrm{PCE}}\left(\hat{Y}, Y\right) &= \frac{1}{|\mathrm{fg}|}\sum_{y \in Y, \hat{y} \in \hat{Y}}{y\log{\left(\hat{y}\right)}} \\
        &+ \frac{1}{|\mathrm{bg}|}\sum_{y \in Y, \hat{y} \in \hat{Y}}{\left(1-y\right)\log{\left(1-\hat{y}\right)}},
    \end{aligned}
    \label{eq:pce_loss}
\end{equation}
where $|\mathrm{fg}|$ is the number of foreground pixels, $|\mathrm{bg}|$ is the number of background pixels, $Y$ is the ground-truth labels, and $\hat{Y}$ are the outputs of dense prediction heads activated by a sigmoid function. This PCE technique is inspired by the influential edge detection method HED \cite{xie2015holistically}.

\setstretch{0.97}

\subsection{Spatial affinity}\label{sec:spatial_affinity}

Spatial affinity is the pixel similarity in terms of visual properties and positions. As point annotations are sparse in nature, they are inadequate to provide sufficient information to determine the shape of the affordance region. Hence, an auxiliary method is in need.

Intuitively, the point annotations of affordance could be propagated to visual regions with similar colors and nearby positions, sharing the same spirit of image segmentations. Modern treatments using \ac{crf} loss~\cite{krahenbuhl2011efficient,tang2018regularized,obukhov2019gated} can fill this gap between point supervision and full supervision. In particular, sparse-connected \ac{crf} loss~\cite{tang2018regularized} has demonstrated similar performance while avoiding the time-consuming process in the dense-connected \ac{crf} loss.

First, we define the kernel as:
\begin{equation}
    \small%
    K_{ab} = \sum_{p=1}^{P}{w^{\left(p\right)} \cdot \exp{\left\{-\frac{1}{2}\left|\frac{f_{a}^{\left(p\right)}-f_{b}^{\left(p\right)}}{\sigma^{\left(p\right)}}\right|^{2}\right\}}},
\end{equation}
where $K_{ab}$ is the weighted sum of $P$ kernels, $w^{\left(p\right)}$ is the weight of kernel $p$, $\sigma^{\left(p\right)}$ is a hyperparameter to control the $\mathrm{p^{th}}$  kernel's bandwidth, and $f_a^{\left(p\right)}$ ($f_b^{\left(p\right)}$) is the value of kernel-specific feature vector at position $a$ ($b$).

Next, we calculate the weighted sum of kernels at the position pair $(a, b)$:
\begin{equation}
    \small%
    \begin{aligned}
        \psi_{a,b}\left(\hat{Y}\right) &= \sum_{i,j \in \left[1,C\right]}{\mu \left(i,j\right)\cdot \hat{y}_a\left(i\right) \cdot \hat{y}_b\left(j\right)\cdot K_{ab}},\\
        \mu \left(i, j\right) &= \left\{\begin{array}{ll}
                    0, \quad{} \text { if } i=j \\
                    1, \quad{} \text { otherwise }
                    \end{array}\right.,
    \end{aligned}
\end{equation}
where $\mu \left(i,j\right)\cdot \hat{y}_a\left(i\right) \cdot \hat{y}_b\left(j\right)$ is a relaxed Potts model of class compatibility, and $C$ is the number of affordance classes in the dataset (Here $C=2$ as we apply the CRF loss on foreground and background separately.).

\begin{figure}[t!]
    \centering
    \includegraphics[width=\linewidth]{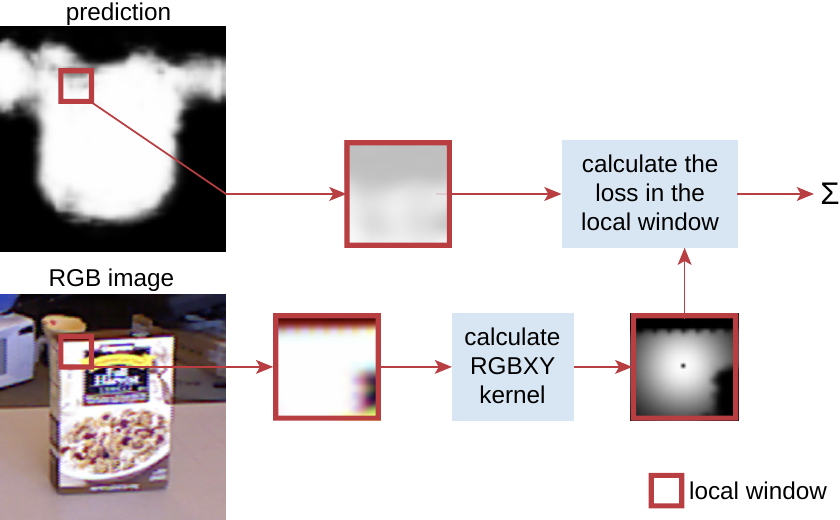}
    \caption{\textbf{Spatial loss by \ac{crf}.} First, a local window slides from the top left to the bottom right to generate a sequence of RGBXY kernels, which are later merged with position information. Next, a similar operation divides the foreground and background predictions into patches, whose size equals the kernels'. Finally, the total loss is the sum of the losses on all local windows.}
    \label{fig:crf}
\end{figure}

Finally, we use an RGBXY kernel to capture the color similarity and position proximity by \ac{crf}:
\begin{equation}
    \small%
    \mathcal{L}_{\mathrm{spatial}} \left(\hat{Y}, X\right) = \frac{1}{N} \sum_{a=1}^{N}{\sum_{b \in \Omega_r\left(a\right) \setminus \left\{a\right\}}{\psi_{a,b}\left(\hat{Y}\right)}},
    \label{eq:spatial_loss}
\end{equation}
where $N$ is the total number of pixels, and $\Omega_r\left(a\right)$ denotes a sliding window at $\left[a_x-r, a_x+r\right] \times \left[a_y-r,a_y+r\right]$; $r$ is the half-width of the slide window. All feature vectors (\ie, RGBXY kernel vectors) are generated by the image $X$, the position matrix, and the binary dense prediction $\hat{Y}$. \cref{fig:crf} illustrates the spatial loss of a sliding window using \ac{crf}.

\subsection{Label affinity}\label{sec:label_affinity}

Label affinity is the structured relation between object labels and affordance labels. We devise the label affinity to model two essential relations: (i) an intra-relation that models the relations between different affordance classes and the relations between different objects, and (ii) an inter-relation that models the relation between affordance and object classes. \cref{fig:architecture}b illustrates this \ac{lai} head.
Similar to the prior design~\cite{nauata2019structured}, we leverage a bidirectional recurrent neural network to model label affinity. The module comprises two label layers: the implicit object layer and the explicit affordance layer. We model the inter-relation across the label layers and the intra-relation in each label layer. Each layer outputs the probability of labels activated by a sigmoid function, and the whole structure is defined as a directed graph.

First, the input feature $x_l$ is fed into label layer $l$,
\begin{equation}
    \small%
    x_l = F_{\mathrm{flatten}} W_{F, l}^\mathrm{T} + b_{F, l},
\end{equation}
where
$l$ is the index of the current layer,
$n_l$ is the number of classes in layer $l$,
$W_{F, l} \in \mathbb{R}^{n_l \times C}$ and $b_{F, l} \in \mathbb{R}^{1 \times n_l}$ are learnable parameters,
and $F_{\mathrm{flatten}} \in \mathbb{R}^{1 \times C}$ is a flattened vector of the Transformer backbone's final feature representation $F \in \mathbb{R}^{\frac{H}{2} \times \frac{W}{2} \times \hat{D}}$.

\setstretch{1}

\begin{figure*}[t!]
    \centering
    \includegraphics[width=\linewidth]{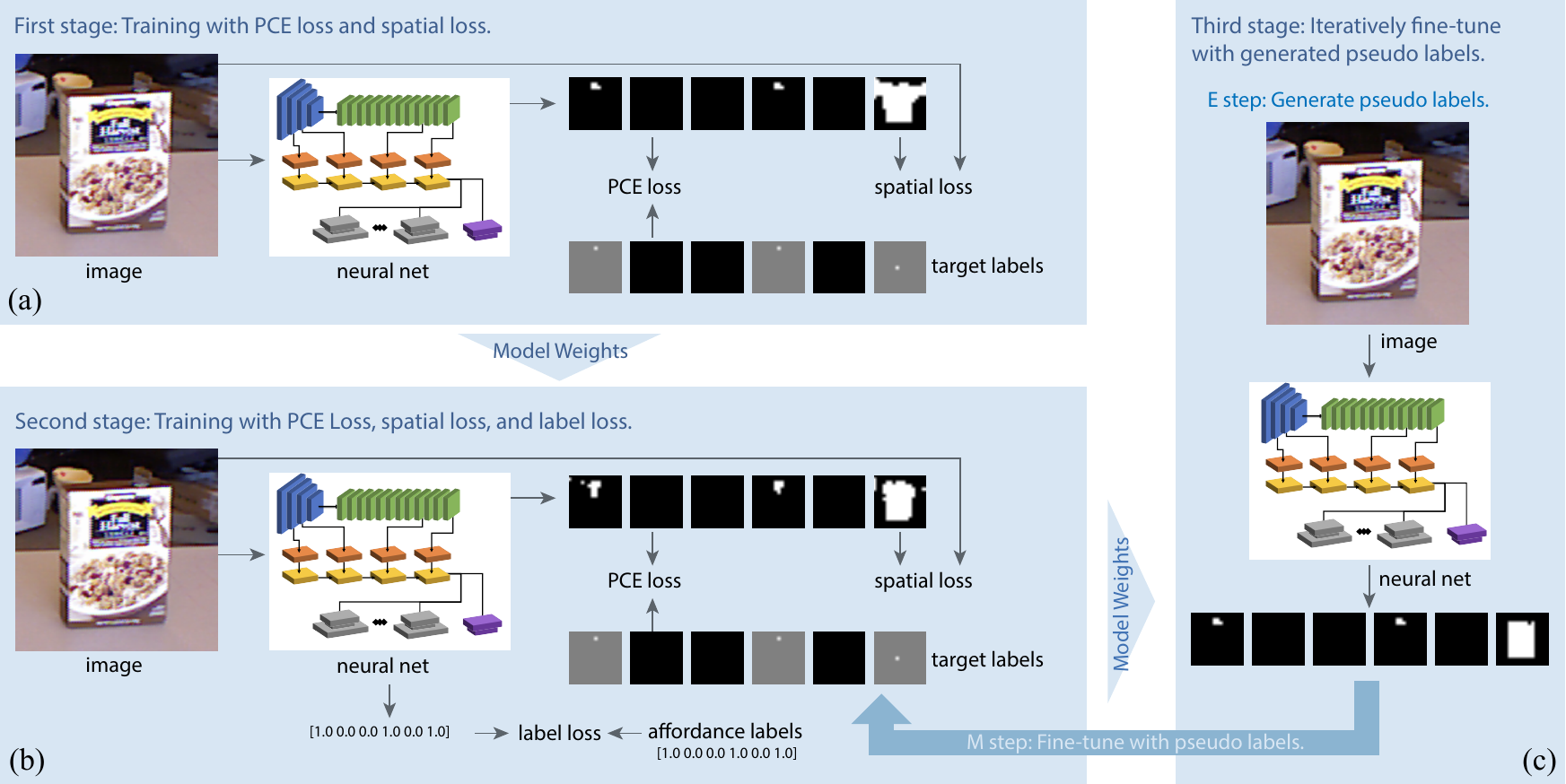}
    \caption{\textbf{Our multi-stage training strategy.} (a) In the first stage, we train our network with the \ac{pce} loss and the spatial loss supervised by point annotations. (b) In the second stage, the \ac{lai} head is added to fine-tune the network by modeling the intra- and inter-relations. This stage is jointly trained by the \ac{pce} loss, the spatial loss, and the label loss. (c) In the third stage, we use the \ac{em} algorithm~\cite{dempster1977maximum} to generate pseudo labels (E step) and fine-tune the network (M step) iteratively.}
    \label{fig:train}
\end{figure*}

\setstretch{0.98}

Next, to output the probabilities of each label, we use a combination of the result of top-down inference $\overrightarrow{f}_l$ and the result of bottom-up inference $\overleftarrow{f}_l$, conceptually illustrated in \cref{fig:architecture}b:
\begin{equation}
    \small%
    \begin{aligned}%
        \overrightarrow{f}_l &= \overrightarrow{f}_{l-1} \cdot \overrightarrow{W}_{\mathrm{inter}, l}^T + x_{l} \cdot \overrightarrow{W}_{\mathrm{intra}, l}^T + \overrightarrow{b}_l,\\
        \overleftarrow{f}_l &= \overleftarrow{f}_{l+1} \cdot \overleftarrow{W}_{\mathrm{inter}, l}^T + x_{l} \cdot \overleftarrow{W}_{\mathrm{intra}, l}^T + \overleftarrow{b}_l,\\
        f_l &= \overrightarrow{f}_{l} \cdot \overrightarrow{W}_l^T + \overleftarrow{f}_{l} \cdot \overleftarrow{W}_l^T + b_l,
    \end{aligned}
\end{equation}
where $W_{\mathrm{inter}}$ denotes the weight of inter-relation, and $W_{\mathrm{intra}}$ denotes the weight of intra-relation. We generate image-level affordance labels $Z \in \mathbb{R}^{n \times 1}$ by point annotations, where $n$ is the number of affordance classes. Hence, we can use a binary cross-entropy loss to optimize this module:
\begin{equation}
    \small%
    \begin{aligned}%
        \mathcal{L}_{\mathrm{label}}\left(\hat{Z}, Z\right) &= \frac{1}{n} \sum_{z \in Z, \hat{z} \in \hat{Z}}{z \cdot \log{\left(\hat{z}\right)}} \\
        & + \frac{1}{n} \sum_{z \in Z, \hat{z} \in \hat{Z}}{\left(1-z\right) \cdot \log{\left(1-\hat{z}\right)}},
    \end{aligned}
    \label{eq:label_loss}
\end{equation}
where $n$ is the number of labels in the last layer (it's equal to the number of affordance classes in our settings), $Z$ is the ground-truth label, and $\hat{Z}$ is the output of the last layer activated by a sigmoid function.

\subsection{Training}

\paragraph{Objective}

Given sparse point annotations, training requires converting them to image-like pseudo labels $Y$ as shown in prior work~\cite{sawatzky2017weakly,sawatzky2017adaptive}:
\begin{equation}
    \small%
    Y\left(i, j\right)=\left\{\begin{array}{ll}
                1, & \text{if } \left(i-p^k_i\right)^2 + \left(j-p^k_j\right)^2 < \sigma_{\mathrm{d}} \\
                0, & \text{otherwise}
                \end{array}\right.,
\end{equation}
where $\left(i, j\right)$ denotes the 2-D position of $Y$, $\left(p^k_i, p^k_j\right)$ denotes the 2-D position of the $k^{\mathrm{th}}$ point annotation, and $\sigma_{\mathrm{d}}$ is a hyperparameter that controls the size of pseudo patterns.
Of note, setting $\sigma_{\mathrm{d}}>1$~\cite{sawatzky2017adaptive} is equivalent to increasing the number of the point annotations, facilitating the feature learning of affordance regions; this introduces an effect akin to the spatial affinity. However, such dilation of point annotations also introduces undesirable artifacts in local regions with intricate boundaries. Hence, in this work, we set $\sigma_{\mathrm{d}}=1\mathrm{px}$ to ensure the correctness of $Y$.

\begin{table*}[t!]
    \centering
    \caption{\textbf{Comparisons between our methods and prior methods on the object split and the actor split.}}
    \label{tab:comparison}
    \setlength{\tabcolsep}{3pt}
    \scalebox{1.0}{%
        \begin{tabular}{@{}clccccccc@{}}
            \toprule
            & methods & openable$\uparrow$ & cuttable$\uparrow$ & pourable$\uparrow$ & containable$\uparrow$ & supportable$\uparrow$ & holdable$\uparrow$ & \acs{metric}$\uparrow$ \\
            \midrule
            \multirow{4}{*}{\rotatebox{90}{object split}}
             & WTP~\cite{bearman2016s} & 0.01 & 0.00 & 0.09 & 0.02 & 0.03 & 0.19 & 0.06 \\
             & Johann Sawatzky \etal.~\cite{sawatzky2017weakly} & 0.11 & 0.09 & 0.21 & 0.28 & 0.36 & 0.56 & 0.27 \\
             & Johann Sawatzky \etal.~\cite{sawatzky2017adaptive} & 0.15 & 0.21 & 0.37 & 0.45 & \textbf{0.61} & 0.54 & 0.39 \\
             & \textbf{Ours} & \textbf{0.52} & \textbf{0.49} & \textbf{0.55} & \textbf{0.57} & 0.57 & \textbf{0.65} & \textbf{0.60} \\
            \midrule
            \multirow{4}{*}{\rotatebox{90}{actor split}}
             & WTP~\cite{bearman2016s} & 0.13 & 0.00 & 0.08 & 0.10 & 0.11 & 0.22 & 0.11 \\
             & Johann Sawatzky \etal.~\cite{sawatzky2017weakly} & 0.23 & \textbf{0.14} & 0.28 & 0.33 & 0.24 & 0.42 & 0.27 \\
             & Johann Sawatzky \etal.~\cite{sawatzky2017adaptive} & \textbf{0.50} & 0.00 & 0.39 & \textbf{0.43} & \textbf{0.64} & 0.56 & 0.42 \\
             & \textbf{Ours} & 0.34 & 0.07 & \textbf{0.40} & 0.40 & 0.49 & \textbf{0.59} & \textbf{0.46} \\
            \bottomrule
        \end{tabular}%
    }%
\end{table*}

\paragraph{Training strategy}

We devise a three-stage training strategy to fully unleash the power of dual affinity modeling; see \cref{fig:train}. Our training strategy is efficient: We directly use $Y$ in the first/second stage and further refine $Y$ without time-consuming post-processing in the third stage.

\setstretch{0.98}

In the first stage, we only use the \ac{pce} loss \cref{eq:pce_loss} and the spatial loss \cref{eq:spatial_loss} to train our network; the total loss is:
\begin{equation}
    \small%
    \mathcal{L}_{\mathrm{s1}} = \lambda_1 \mathcal{L}_{\mathrm{PCE}} + \lambda_2 \mathcal{L}_{\mathrm{spatial}},
\end{equation}
where $\lambda_1 = 1$, and $\lambda_2 = 0.1$ in our settings.

In the second stage, we add the \ac{lai} head to exploit the inter-/intra-relations and only use the image-level affordance labels \cref{eq:label_loss} to supervise the \ac{lai} head. The total loss is:
\begin{equation}
    \small%
    \mathcal{L}_{\mathrm{s2}} = \lambda_1 \mathcal{L}_{\mathrm{PCE}} + \lambda_2 \mathcal{L}_{\mathrm{spatial}} + \lambda_3 \mathcal{L}_{\mathrm{label}},
    \label{eq:total_loss}
\end{equation}
where $\lambda_1 = 1$, $\lambda_2 = 0.1$, and $\lambda_3 = 1$ in our settings. This strategy eases the training of LAI head.

In the third stage, we adopt the \acf{em} algorithm to boost the network performance progressively without any post-processing trick to refine pseudo labels (\eg, dense \ac{crf} inference~\cite{krahenbuhl2011efficient}).
The iteration of \ac{em} includes two steps: (i) In the E step, we freeze our network to predict the affordance maps and use the flood-fill algorithm to extract all foreground regions annotated by sparse point annotations. (ii) In the M step, we train our network supervised by these pseudo labels and affordance labels, with the same total loss as in \cref{eq:total_loss}.

\paragraph{Flood-fill algorithm}

Flood fill is an algorithm that propagates the connected and similarly-colored region from the initial nodes; \ie, the point annotations in our settings. We only use the connectivity for flooding since the affordance segmentation yields a set of binary maps. There are two common strategies to find eligible adjacent nodes: four-way (in the shape of a cross) and eight-way (in the shape of a rectangle). The implementation of the flood-fill algorithm is usually by a stack- or a queue-based search, wherein the procedure is similar to the depth-first search in a graph. \cref{alg:flood_fill} sketches the pseudocode of the flood-fill algorithm.

\setstretch{1}

\begin{algorithm}[ht!]
    \small
    \caption{\textsc{Flood-fill Algorithm}}\label{alg:flood_fill}
    \SetKwFunction{push}{push}
    \SetKwFunction{pop}{pop}
    \SetKwFunction{neighbor}{neighbor}
    \KwIn{
    $p$: one of the sparse point annotations \newline
    $\hat{Y}$: the binary prediction
    }
    \KwOut{
    $Y$: the processed binary prediction
    }
    \KwData{
    $Q$: an empty queue
    }
    \Begin{
    $Y \leftarrow \mathbf{0}$\;
    $Q \leftarrow \push{p}$\;
        \While{$Q$ is not empty}{
            $p_\mathrm{new} \leftarrow$ \pop{$Q$}\;
            \If{$\hat{Y}({p_\mathrm{new}}) = 0$}
            {
                continue\;
            }
            $Y(p_\mathrm{new}) \leftarrow 1$\;
            \ForEach{$p_\mathrm{neighbor}$ in \neighbor{$p_\mathrm{new}$}}{
                $Q \leftarrow $ \push{$p_\mathrm{neighbor}$}\;
            }
        }
    }%
\end{algorithm}%

\setstretch{1}
\section{Experiments}

We evaluate our methods on the CAD120\footnote{dataset link: \url{https://zenodo.org/record/495570}} dataset~\cite{koppula2013learning}, which contains two splits: (i) Object split, with no same central object class between the training and the test data. (ii) Actor split, with no same video actor between the training and the test data. Note that the actor split is much more difficult than the object split for all methods inspected, because of scene context domain gap. For evaluation metrics, we adopt the \ac{metric} to compare our method's performance with prior methods.

\paragraph{Implementation details}

On each dataset split, we train our network with a batch size of 4 and an initial learning rate $\mathrm{lr}_{\mathrm{init}}$ of 0.001. We use the function $\mathrm{lr} = \mathrm{lr}_{\mathrm{init}} \times \left(1 - \frac{\mathrm{epoch}}{100}\right)^{0.9}$ to decay the learning rate during training, with an SGD optimizer with a momentum of 0.9 and a weight decay of 0.0001.
In the first and the second stages, given the sparse point supervision (in fact, one point per annotation), we draw disks (\ie, set $\sigma_{\mathrm{d}}=1$px) centered on the point annotation positions to generate pseudo labels, which are used to train our network. In the third stage, we use the predictions in the E step to update the pseudo labels. Every stage lasts 100 epochs. Besides, we apply the EM algorithm every 10 epochs in the third stage, accounting to 10 EM rounds.

\begin{figure*}[t!]
    \centering
    \hfill%
    \begin{subfigure}[t]{0.46\linewidth}
        \centering
        \begin{overpic}
            [width=0.2\linewidth]{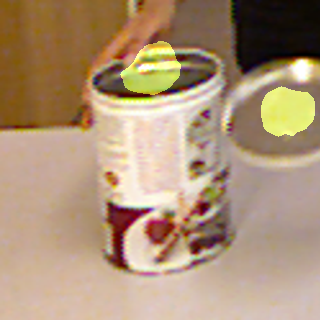}%
            \put(0,105){\small\color{black}{only \ac{pce} loss}}
            \put(-20,12){\rotatebox{90}{\color{black}{openable}}}
        \end{overpic}%
        \begin{overpic}
            [width=0.2\linewidth]{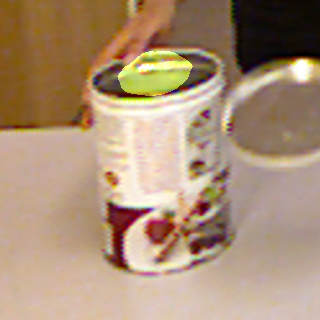}%
            \put(15,105){\small\color{black}{first stage}}
        \end{overpic}%
        \begin{overpic}
            [width=0.2\linewidth]{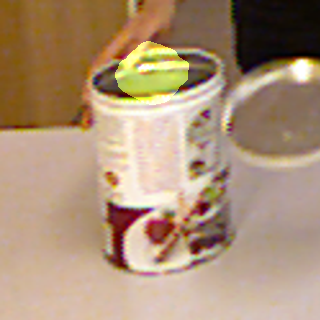}%
            \put(2,105){\small\color{black}{second stage}}
        \end{overpic}%
        \begin{overpic}
            [width=0.2\linewidth]{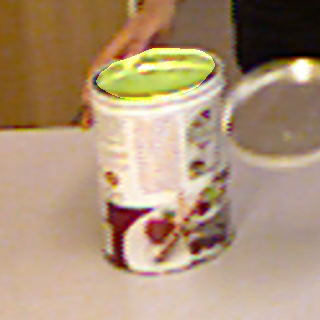}%
            \put(15,105){\small\color{black}{third stage}}
        \end{overpic}%
        \begin{overpic}
            [width=0.2\linewidth]{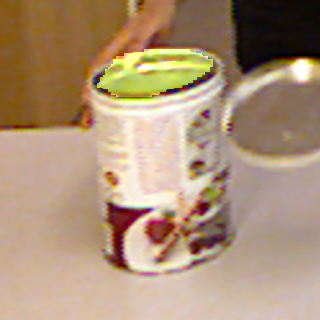}%
            \put(2,105){\small\color{black}{ground truth}}
        \end{overpic}%
        \\%
        \begin{overpic}
            [width=0.2\linewidth]{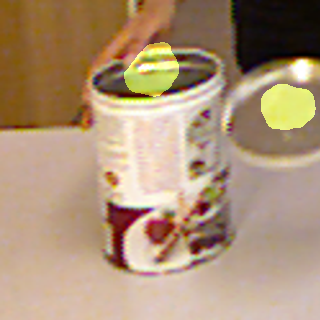}%
            \put(-20,2){\rotatebox{90}{\color{black}{containable}}}
        \end{overpic}%
        \includegraphics[width=0.2\linewidth]{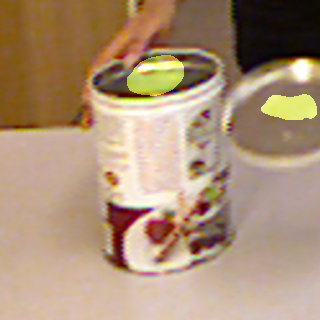}%
        \includegraphics[width=0.2\linewidth]{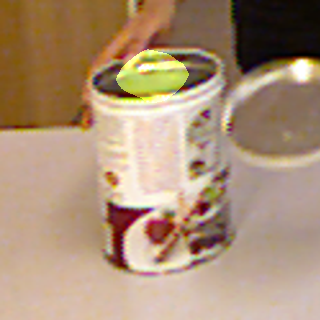}%
        \includegraphics[width=0.2\linewidth]{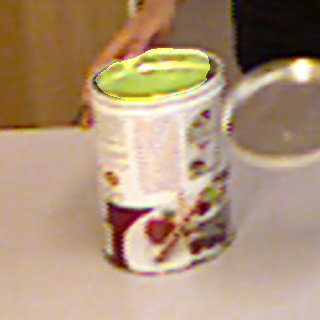}%
        \includegraphics[width=0.2\linewidth]{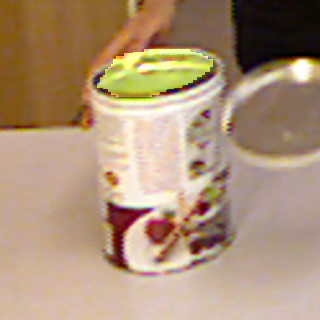}%
        \\%
        \begin{overpic}
            [width=0.2\linewidth]{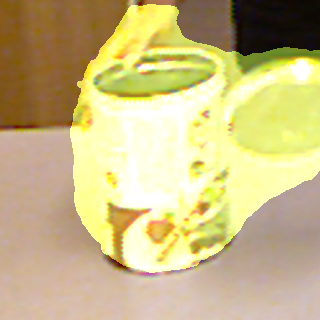}%
            \put(-20,15){\rotatebox{90}{\color{black}{holdable}}}
        \end{overpic}%
        \includegraphics[width=0.2\linewidth]{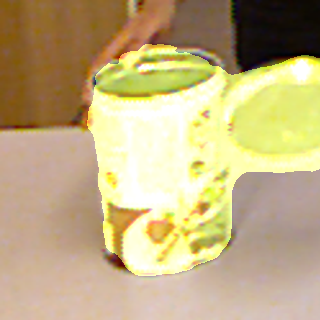}%
        \includegraphics[width=0.2\linewidth]{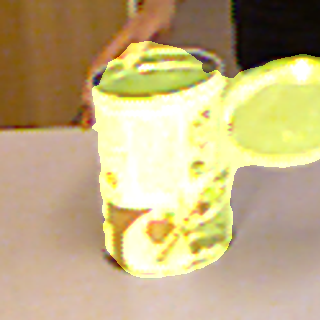}%
        \includegraphics[width=0.2\linewidth]{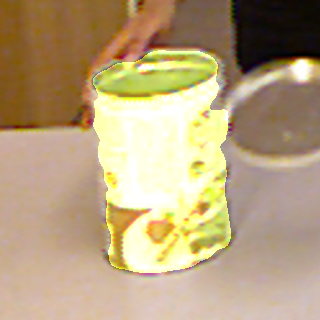}%
        \includegraphics[width=0.2\linewidth]{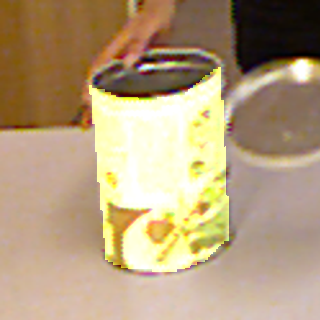}%
        \caption{}
        \label{fig:qualitative:q1}
    \end{subfigure}%
    \hfill%
    \begin{subfigure}[t]{0.46\linewidth}
        \centering
        \begin{overpic}
            [width=0.2\linewidth]{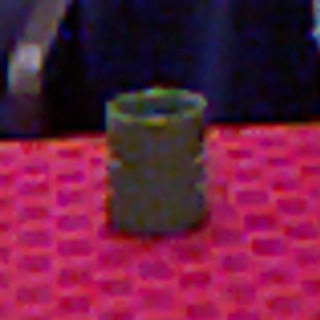}%
            \put(0,105){\small\color{black}{only \ac{pce} loss}}
            \put(-20,12){\rotatebox{90}{\color{black}{pourable}}}
        \end{overpic}%
        \begin{overpic}
            [width=0.2\linewidth]{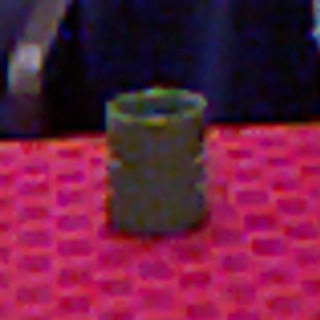}%
            \put(15,105){\small\color{black}{first stage}}
        \end{overpic}%
        \begin{overpic}
            [width=0.2\linewidth]{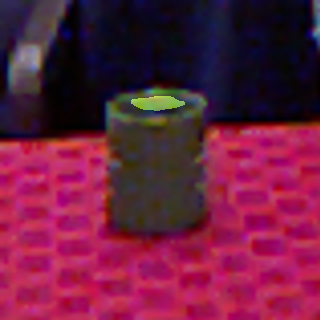}%
            \put(2,105){\small\color{black}{second stage}}
        \end{overpic}%
        \begin{overpic}
            [width=0.2\linewidth]{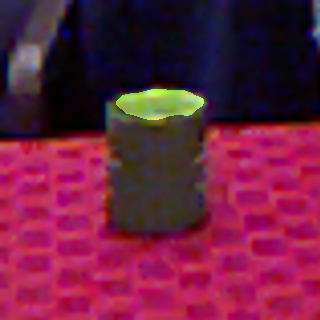}%
            \put(15,105){\small\color{black}{third stage}}
        \end{overpic}%
        \begin{overpic}
            [width=0.2\linewidth]{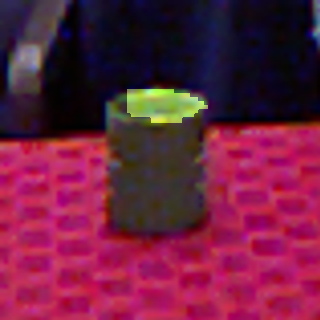}%
            \put(2,105){\small\color{black}{ground truth}}
        \end{overpic}%
        \\%
        \begin{overpic}
            [width=0.2\linewidth]{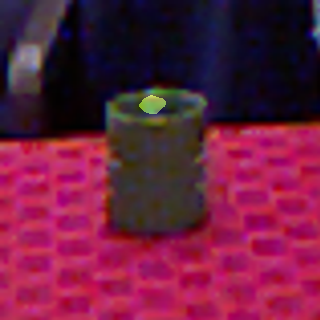}%
            \put(-20,2){\rotatebox{90}{\color{black}{containable}}}
        \end{overpic}%
        \includegraphics[width=0.2\linewidth]{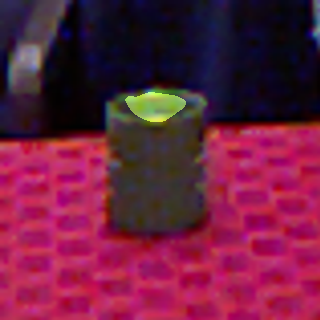}%
        \includegraphics[width=0.2\linewidth]{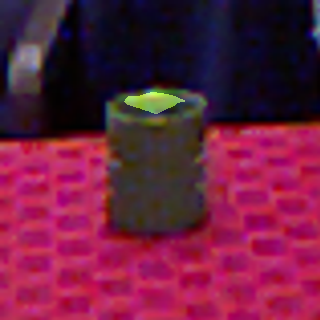}%
        \includegraphics[width=0.2\linewidth]{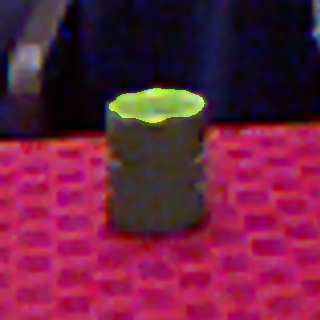}%
        \includegraphics[width=0.2\linewidth]{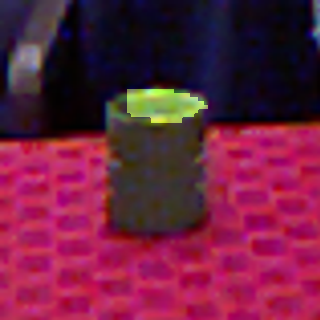}%
        \\%
        \begin{overpic}
            [width=0.2\linewidth]{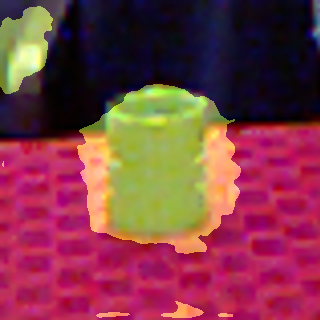}%
            \put(-20,15){\rotatebox{90}{\color{black}{holdable}}}
        \end{overpic}%
        \includegraphics[width=0.2\linewidth]{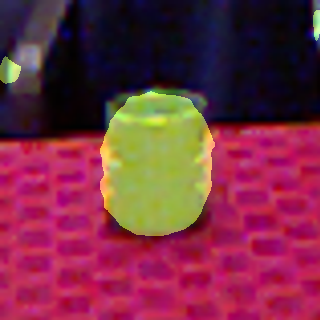}%
        \includegraphics[width=0.2\linewidth]{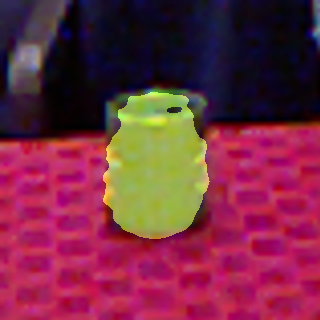}%
        \includegraphics[width=0.2\linewidth]{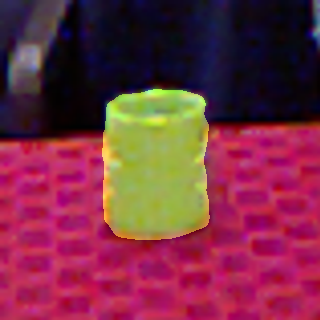}%
        \includegraphics[width=0.2\linewidth]{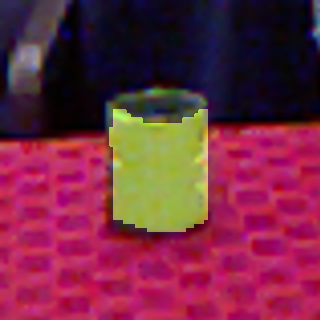}%
        \caption{}
        \label{fig:qualitative:q2}
    \end{subfigure}%
    \\%
    \hfill%
    \begin{subfigure}[t]{0.46\linewidth}
        \centering
        \begin{overpic}
            [width=0.2\linewidth]{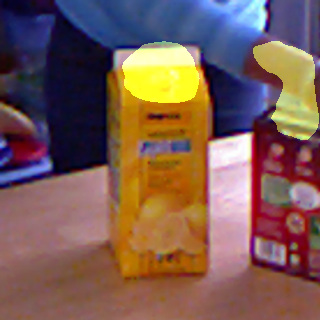}%
            \put(-20,12){\rotatebox{90}{\color{black}{openable}}}
        \end{overpic}%
        \includegraphics[width=0.2\linewidth]{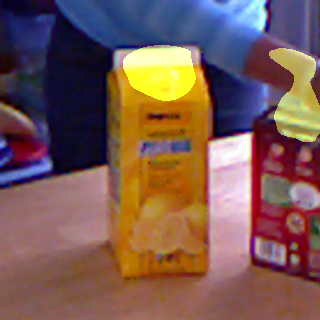}%
        \includegraphics[width=0.2\linewidth]{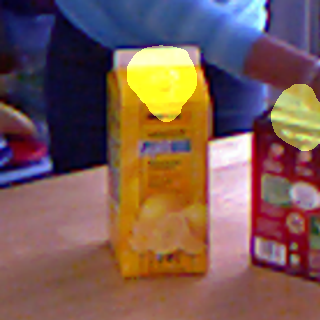}%
        \includegraphics[width=0.2\linewidth]{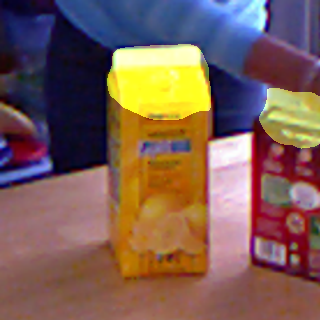}%
        \includegraphics[width=0.2\linewidth]{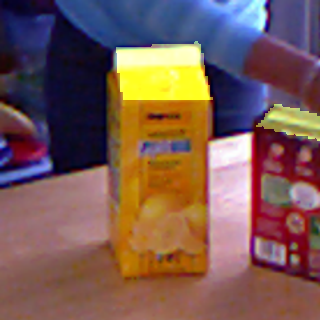}%
        \\%
        \begin{overpic}
            [width=0.2\linewidth]{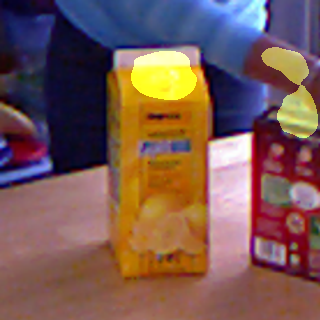}%
            \put(-20,2){\rotatebox{90}{\color{black}{containable}}}
        \end{overpic}%
        \includegraphics[width=0.2\linewidth]{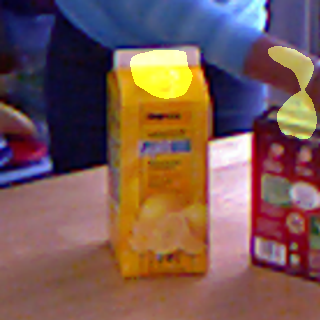}%
        \includegraphics[width=0.2\linewidth]{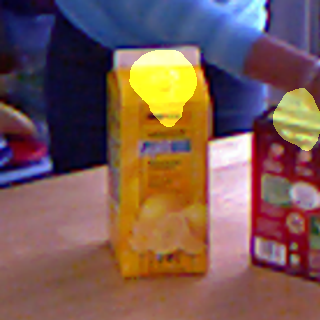}%
        \includegraphics[width=0.2\linewidth]{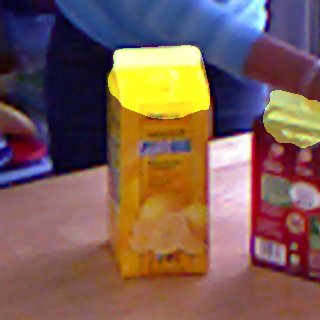}%
        \includegraphics[width=0.2\linewidth]{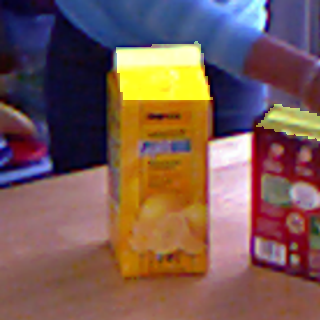}%
        \\%
        \begin{overpic}
            [width=0.2\linewidth]{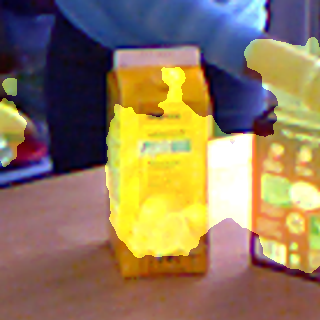}%
            \put(-20,15){\rotatebox{90}{\color{black}{holdable}}}
        \end{overpic}%
        \includegraphics[width=0.2\linewidth]{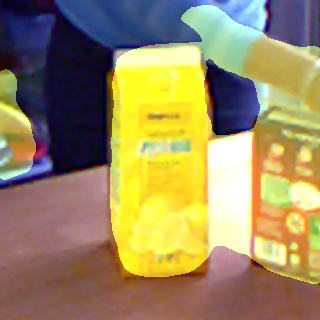}%
        \includegraphics[width=0.2\linewidth]{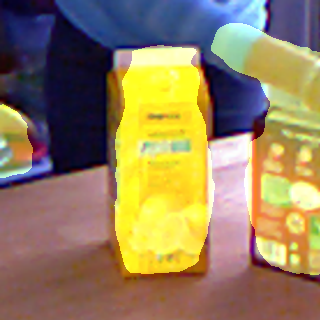}%
        \includegraphics[width=0.2\linewidth]{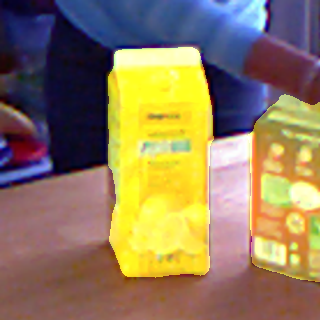}%
        \includegraphics[width=0.2\linewidth]{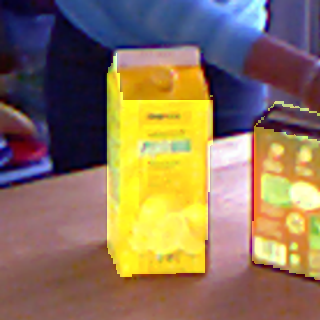}%
        \caption{}
        \label{fig:qualitative:q3}
    \end{subfigure}%
    \hfill
    \begin{subfigure}[t]{0.46\linewidth}
        \centering
        \begin{overpic}
            [width=0.2\linewidth]{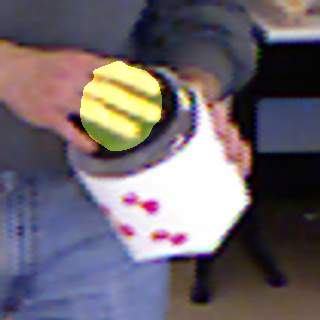}%
            \put(-20,12){\rotatebox{90}{\color{black}{openable}}}
        \end{overpic}%
        \includegraphics[width=0.2\linewidth]{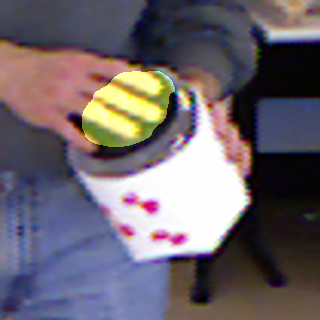}%
        \includegraphics[width=0.2\linewidth]{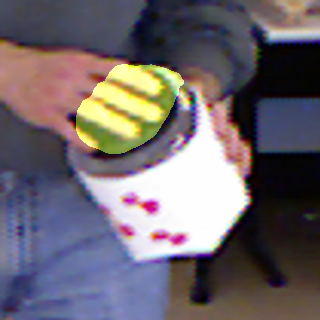}%
        \includegraphics[width=0.2\linewidth]{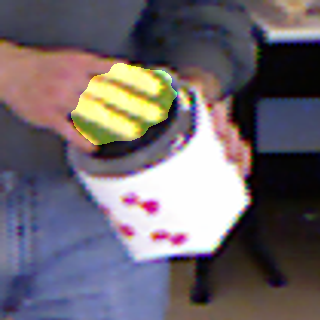}%
        \includegraphics[width=0.2\linewidth]{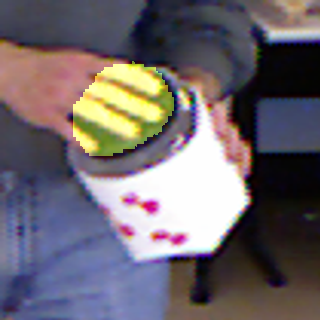}%
        \\%
        \begin{overpic}
            [width=0.2\linewidth]{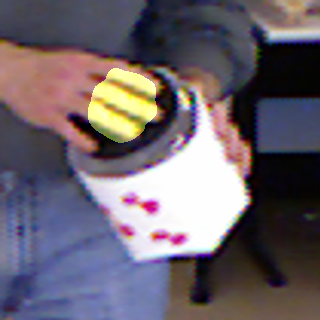}%
            \put(-20,2){\rotatebox{90}{\color{black}{containable}}}
        \end{overpic}%
        \includegraphics[width=0.2\linewidth]{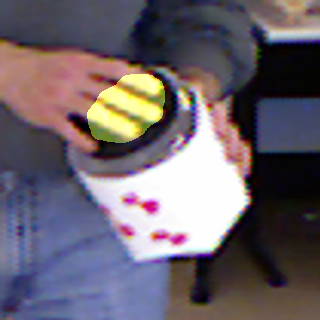}%
        \includegraphics[width=0.2\linewidth]{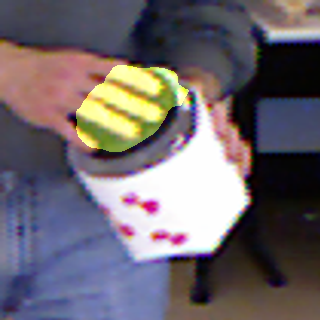}%
        \includegraphics[width=0.2\linewidth]{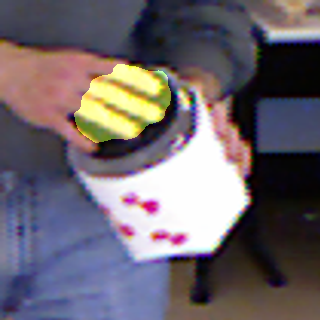}%
        \includegraphics[width=0.2\linewidth]{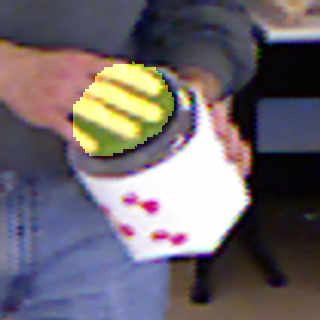}%
        \\%
        \begin{overpic}
            [width=0.2\linewidth]{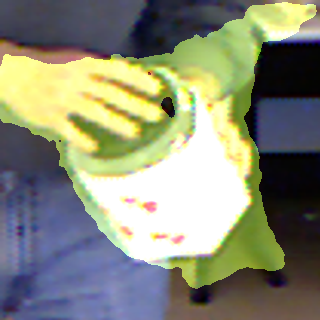}%
            \put(-20,15){\rotatebox{90}{\color{black}{holdable}}}
        \end{overpic}%
        \includegraphics[width=0.2\linewidth]{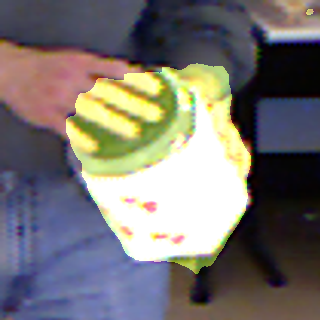}%
        \includegraphics[width=0.2\linewidth]{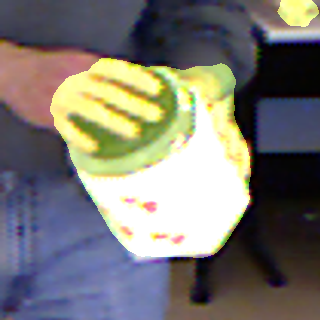}%
        \includegraphics[width=0.2\linewidth]{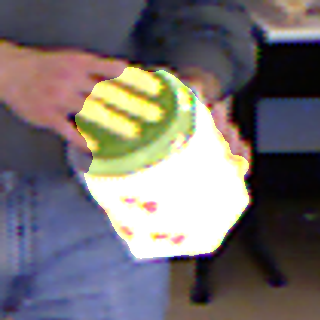}%
        \includegraphics[width=0.2\linewidth]{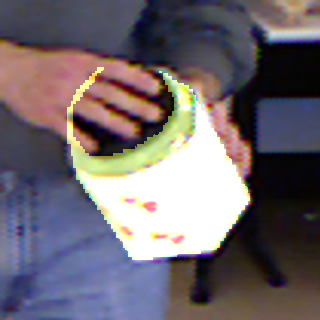}%
        \caption{}
        \label{fig:qualitative:q4}
    \end{subfigure}%
    \caption{\textbf{Qualitative results in three different affordance classes.}}
    \label{fig:qualitative}
\end{figure*}

\begin{table*}[b!]
    \centering
    \caption{\textbf{Ablations of our methods on the object spit and the actor split.}}
    \label{tab:ablation}
    \setlength{\tabcolsep}{3pt}
    \scalebox{0.8}{%
        \begin{tabular}{@{}cccccccccccc@{}}
            \toprule
             & \ac{pce} & spatial affinity & label affinity & \ac{em} & openable$\uparrow$ & cuttable$\uparrow$ & pourable$\uparrow$ & containable$\uparrow$ & supportable$\uparrow$ & holdable$\uparrow$ & \acs{metric}$\uparrow$ \\
            \midrule
            \multirow{4}{*}{\rotatebox{90}{object split}} & \checkmark & & & & 0.39 & \textbf{0.59} & 0.33 & 0.37 & \textbf{0.64} & 0.38 & 0.41 \\
             & \checkmark & \checkmark & & & 0.44 & 0.45 & 0.49 & 0.50 & 0.57 & 0.48 & 0.49 \\
             & \checkmark & \checkmark & \checkmark & & \textbf{0.55} & 0.49 & 0.50 & 0.55 & 0.58 & 0.52 & 0.53 \\
             & \checkmark & \checkmark & \checkmark & \checkmark & 0.52 & 0.49 & \textbf{0.55} & \textbf{0.57} & 0.57 & \textbf{0.65} & \textbf{0.60} \\
            \midrule
            \multirow{4}{*}{\rotatebox{90}{actor split}} & \checkmark & & & & 0.27 & 0.05 & 0.37 & 0.34 & \textbf{0.50} & 0.37 & 0.37 \\
             & \checkmark & \checkmark & & & 0.21 & 0.01 & 0.41 & 0.35 & 0.48 & 0.47 & 0.39 \\
             & \checkmark & \checkmark & \checkmark & & 0.26 & 0.00 & \textbf{0.45} & 0.39 & 0.48 & 0.55 & 0.44 \\
             & \checkmark & \checkmark & \checkmark & \checkmark & \textbf{0.34} & \textbf{0.07} & 0.40 & \textbf{0.40} & 0.49 & \textbf{0.59} & \textbf{0.46} \\
            \bottomrule
        \end{tabular}%
    }%
\end{table*}

\paragraph{Quantitative results}

Our methods outperform the state-of-the-art on both the object split and the actor split in terms of the \ac{metric} by 53.8\% and 9.5\%, relatively; see \cref{tab:comparison}.
In particular, our method significantly outperforms prior methods in individual affordance classes (except ``supportable'') on object split. It is on par with the state-of-the-art methods in individual affordance classes on actor split.

\paragraph{Qualitative results}

We visualize four examples as the qualitative results in \cref{fig:qualitative}. Below, we summarize some notable observations in each example.

\begin{itemize}[leftmargin=*,noitemsep,nolistsep]
    \item \cref{fig:qualitative:q1}: The \textit{containable} region on the lid is eradicated after the second stage, indicating that our methods can restrain the prediction of the unreasonable region when using the label affinity loss in the second stage.
    
    \item \cref{fig:qualitative:q2}: The \textit{pourable} region on the mug rim emerges after introducing the label affinity. This result demonstrates the efficacy of our design of label affinity loss.

    \item \cref{fig:qualitative:q3}: Distinguishing different affordance regions with only one point per affordance region is challenging. The spatial affinity loss cannot solve this difficulty, as shown by the prediction of \textit{holdable} region in the first stage. Again, this challenge is solved by the label affinity loss, which models the intra-relation of object labels.
    
    \item \cref{fig:qualitative:q4}: Despite being annotated with other affordance classes, the top-right region lacks the point annotation for \textit{holdable}. Hence, the algorithm struggles in prediction with only \ac{pce} loss and the spatial affinity loss. In fact, this label is also challenging for humans due to its small scale, color distortion, and incomplete shape. Yet, our network still discovers the top-right \textit{holdable} region by introducing the label affinity loss in the second stage.
\end{itemize}

\paragraph{Ablations of loss functions}

We further conduct some ablations of loss functions in our method, summarized in \cref{tab:ablation}. 
On object split, we observe a 19.5\% performance improvement in \ac{metric} when adding the spatial affinity loss and another 8.2\% in \ac{metric} when adding the label affinity loss. These improvements demonstrate that both spatial affinity and label affinity play a central role in filling the gap between full and point supervision. Moreover, without time-consuming post-processing, the \ac{em} algorithm brings in a 13.2\% performance improvement in \ac{metric}, demonstrating the efficacy of our design.
Although we did not observe similar significant improvements on actor split, our design still yields a similar performance or progressively improved performance with additional modules added.

\paragraph{Ablations of $\mathbf{\sigma_{\mathrm{d}}}$}

\begin{figure}[t!]
    \centering
    \includegraphics[width=0.9\linewidth]{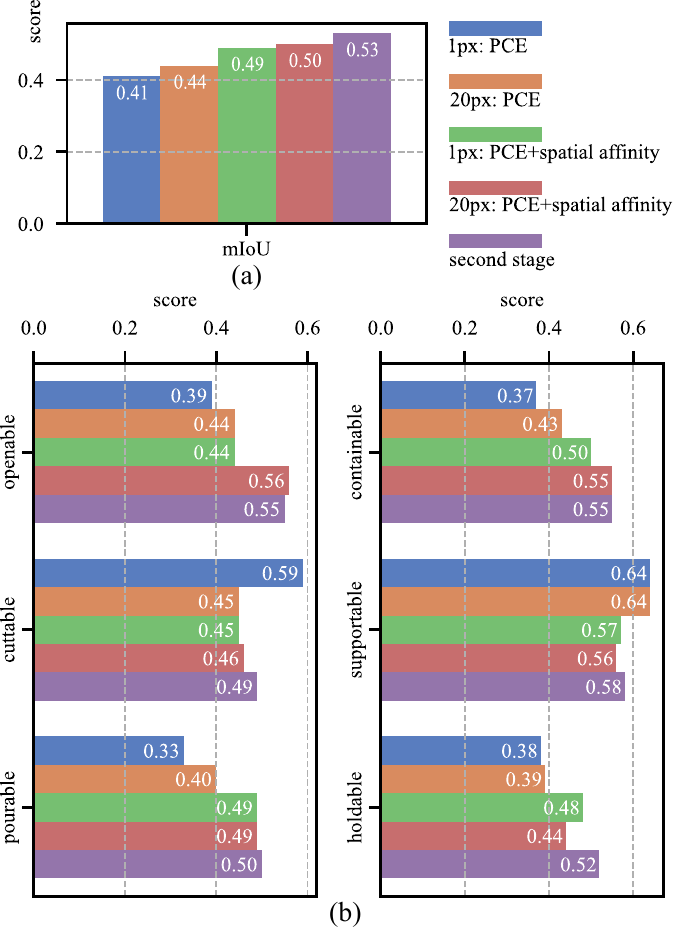}
    \caption{\textbf{Ablations of $\mathbf{\sigma_{\mathrm{d}}}$.} We experiment and compare five settings: (i) \ac{pce} loss with $\sigma_{\mathrm{d}} = 1\mathrm{px}$, (ii) \ac{pce} loss and \ac{crf} loss with $\sigma_{\mathrm{d}} = 1\mathrm{px}$ (\ie, the first stage), (iii) \ac{pce} loss with $\sigma_{\mathrm{d}} = 20\mathrm{px}$, (iv) \ac{pce} loss and \ac{crf} loss with $\sigma_{\mathrm{d}} = 20\mathrm{px}$, and (v) the performance obtained up to the second stage in training.}
    \label{fig:ablation_diameter}
\end{figure}

We further ablate a crucial parameter $\sigma_{\mathrm{d}}$. Prior methods suggest that a larger $\sigma_{\mathrm{d}}$ may lead to better performance~\cite{sawatzky2017adaptive} ($\sigma_{\mathrm{d}} = 20\mathrm{px}$). At first glance, this empirical result is echoed by the comparison shown in \cref{fig:ablation_diameter}a. Specifically, we observe that $\sigma_{\mathrm{d}} = 20\mathrm{px}$ yields better \ac{metric} compared to $\sigma_{\mathrm{d}} = 1\mathrm{px}$ (i) with \ac{pce} loss only (the orange \vs the blue) and (ii) with both \ac{pce} loss and spatial affinity loss (the red \vs the green). However, our further analysis of individual affordance classes refutes the empirical finding that larger $\sigma_{\mathrm{d}}$ may lead to better performance.

\cref{fig:ablation_diameter}b summarizes the results of individual affordance classes, which presents a mixed picture. For instance, when only using the \ac{pce} loss, the \ac{metric} of $\sigma_{\mathrm{d}} = 20\mathrm{px}$ is 7.3\% higher than $\sigma_{\mathrm{d}} = 1\mathrm{px}$; conversely, adding the spatial affinity loss reduces this gap to 2.0\%. Overall, only the affordance classes of \textit{openable}, \textit{pourable}, and \textit{containable} indicate $\sigma_{\mathrm{d}} = 20\mathrm{px}$ is suitable, whereas the others do not. Interestingly, our proposed \ac{lai} head (\ie, the second stage) performs the best among all five conditions, indicating the efficacy of the dual affinity design.

\begin{table}[t!]
    \centering
    \small
    \caption{\textbf{The cumulative percentages of a specific $\mathbf{\sigma_{\mathrm{dm}}}$.}}
    \label{tab:cumulative}
    \begin{tabular}[8pt]{@{}lrr@{}}
        \toprule
        affordance class & $<$ 20px & $<$ 50px \\
        \midrule
        openable & 14.07\% & 66.24\% \\
        cuttable & 57.10\% & 100.00\% \\
        pourable & 8.38\% & 65.54\% \\
        containable & 12.42\% & 70.77\% \\
        supportable & 18.21\% & 61.38\% \\
        holdable & 28.09\% & 67.66\% \\
        \bottomrule
    \end{tabular}
\end{table}

\begin{figure}[t!]
    \centering
    \includegraphics[width=0.7\linewidth]{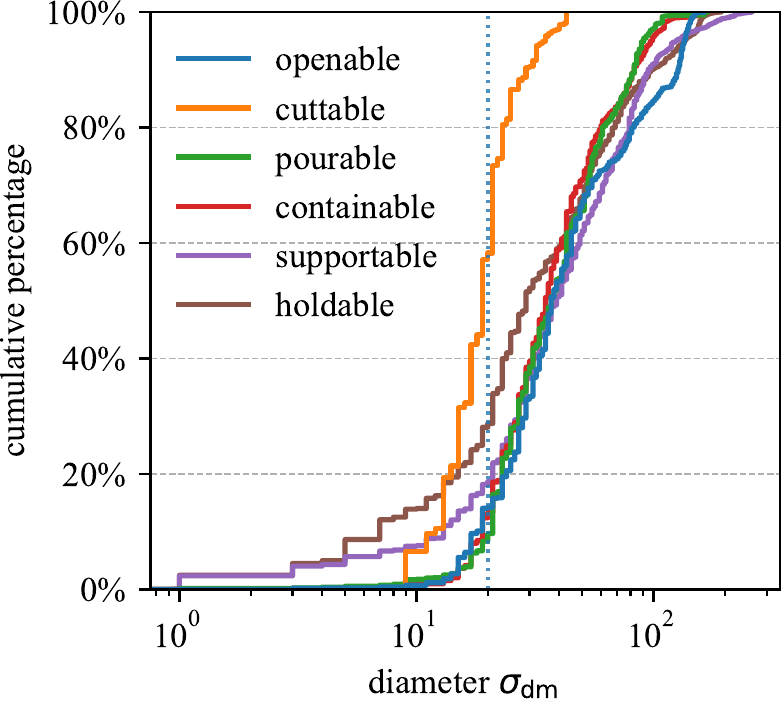}
    \caption{\textbf{Cumulative percentage curve of $\mathbf{\sigma_{\mathrm{dm}}}$.} We compute the appropriate diameter $\sigma_{\mathrm{dm}}$ for each affordance region, defined as the maximal of $\sigma_{\mathrm{d}}$ that ensures the disk generated by it does not exceed the boundary of ground truth.}
    \label{fig:diameter_cdf}
\end{figure}

We further calculate the cumulative percentage of appropriate diameter $\sigma_{\mathrm{dm}}$ for each affordance class; see \cref{fig:diameter_cdf}. Of note, we define the appropriate diameter $\sigma_{\mathrm{dm}}$ for a specific affordance region as the maximal $\sigma_{\mathrm{d}}$, which ensures that the disk drawn by $\sigma_{\mathrm{dm}}$ does not exceed the boundary of the ground-truth region. The cumulative percentage curve accounts for the advantage of $\sigma_{\mathrm{d}} = 20\mathrm{px}$ in \textit{openable}, \textit{pourable}, and \textit{containable} since almost all of $\sigma_{\mathrm{dm}}$ is larger than 20px except \textit{holdable} and \textit{cuttable}; see also \cref{tab:cumulative}. This result echoes our above analysis of \cref{fig:ablation_diameter} and confirms that our setting of $\sigma_{\mathrm{d}} = 1\mathrm{px}$ is a more practical solution.

Taken together, our ablations of $\mathbf{\sigma_{\mathrm{d}}}$ challenges the prior findings~\cite{sawatzky2017adaptive}; instead, our experimental results indicate an optimal solution (\ie, $\sigma_{\mathrm{d}} = 1\mathrm{px}$) is the best. This contradiction calls for future investments in affordance segmentation and related tasks.

\paragraph{Drawback}

The qualitative results in the third stage also reveal a drawback in terms of the generalization of the \ac{em} algorithm. Let us take the \textit{holdable} in \cref{fig:qualitative:q1,fig:qualitative:q3} as an example. When learning an affordance region without any point annotation (either the unlabelled class in \cref{fig:qualitative:q1} or the missing annotation in the central figure's top right corner in \cref{fig:qualitative:q3}), the \ac{em} algorithm removes this region iteratively. As such, despite the final result matching the ground-truth label, the \ac{em} algorithm prevents the method from compensating for incomplete ground-truth annotations.
\section{Conclusion}

In this work, we focused on affordance segmentation with sparse point supervision and proposed a novel Transformer-based dense prediction network architecture with a dual affinity design. Experiments demonstrate that our method achieves state-of-the-art performance on the challenging CAD120 dataset.

Computationally, our analysis reveals three essential ingredients: (i) Our method can explicitly understand the shape of the affordance region by \textbf{spatial affinity}. (ii) \textbf{Label affinity} implicitly helps to distinguish the class boundary of different affordance regions and discovers other affordance regions even without point annotations. (iii) Filtering the affordance regions by flood-fill guided by point annotations in the \textbf{\ac{em} algorithm} can further boost the performance.
   
{\small
\balance
\bibliographystyle{ieee_fullname}
\bibliography{reference}
}

\pagebreak
\appendix

\section{EM Algorithm}\label{sec:em_algo}

Given an input image $x$ sampled from an underlying distribution, we can use 
the network, parameterized by $\theta$, to predict the corresponding affordance segmentation $\hat{y}$. The latent code is defined as the pseudo labels $z$ that have a latent relationship with the ground-truth labels $y$. The log-likelihood objective is defined as:

\begin{equation}
    \mathcal{L}\left(\theta\right) = {\log{p\left(x\mid\theta\right)}}
\end{equation}

To solve this maximum likelihood estimation (MLE) problem, we use the EM algorithm by iteratively applying these two steps:
\begin{itemize}[leftmargin=*,noitemsep,nolistsep]
    \item Expectation (E) step: In the $t$-th E step, we calculate the quantity of $z^{(t)}$ by the following equation:
    \begin{equation}
    \begin{aligned}
        q\left(\theta\mid\theta^{(t)}\right)
        &= \mathrm{E}_{z\sim p\left(\cdot\mid x, \theta^{(t)}\right)}\left[\log{p\left(x, z\mid\theta\right)}\right] \\
        &= \sum_{z\in Z}{p\left(z\mid x, \theta^{(t)}\right)}\log{p\left(x, z\mid\theta\right)}
    \end{aligned}
    \end{equation}
    where $Z$ represents the distribution of the latent code $z$.
    
    \item Maximization (M) step: In the $t$-th M step, we use the stochastic gradient descent (SGD) algorithm to find best $\theta$:
    \begin{equation}
        \theta^{(t+1)} 
        = \mathrm{argmax}_{\theta}{q\left(\theta\mid\theta^{(t)}\right)}
    \end{equation}
\end{itemize}

We provide a brief proof of correctness as follows:

\begin{equation}
\label{eqa:proof}
\begin{aligned}
    &\mathcal{L}\left(\theta^{(t+1)}\right) - \mathcal{L}\left(\theta^{(t)}\right) \\
    = &\log{p\left(x\mid\theta^{(t+1)}\right)} - \log{p\left(x\mid\theta^{(t)}\right)} \\
    = &\sum_{z\in Z}{p\left(z\mid x, \theta^{(t)}\right)\left(\log{\frac{p\left(x,z\mid\theta^{(t+1)}\right)}{p\left(z\mid x,\theta^{(t+1)}\right)}} - \log{\frac{p\left(x,z\mid\theta^{(t)}\right)}{p\left(z\mid x,\theta^{(t)}\right)}}\right)} \\
    = &q\left(\theta^{(t+1)}\mid\theta^{(t)}\right) - q\left(\theta^{(t)}\mid\theta^{(t)}\right) \\
    &- \sum_{z\in Z}{p\left(z\mid x,\theta^{(t)}\right) \log{\frac{p\left(z\mid x,\theta^{(t+1)}\right)}{p\left(z\mid x,\theta^{(t)}\right)}}}
\end{aligned}
\end{equation}

For the first term in \cref{eqa:proof}, since we choose $\theta$ to improve $q\left(\theta\mid\theta^{(t)}\right)$, we can conclude that:

\begin{equation}
    q\left(\theta\mid\theta^{(t+1)}\right) \ge q\left(\theta\mid\theta^{(t)}\right)
\end{equation}

For the last term in \cref{eqa:proof}, we use Jensen’s inequality to get the upper bound of it:

\begin{equation}
\begin{aligned}
    &\sum_{z\in Z}{p\left(z\mid x,\theta^{(t)}\right) \log{\frac{p\left(z\mid x,\theta^{(t+1)}\right)}{p\left(z\mid x,\theta^{(t)}\right)}}} \\
    \leq &\log{\sum_{z\in Z}{p\left(z\mid x,\theta^{(t)}\right)\frac{p\left(z\mid x,\theta^{(t+1)}\right)}{p\left(z\mid x,\theta^{(t)}\right)}}} \\
    = &\log{\sum_{z\in Z}p\left(z\mid x,\theta^{(t+1)}\right)} \\
    = &0
\end{aligned}    
\end{equation}

Thus, \cref{eqa:proof} can be transformed into the following inequation:

\begin{equation}
    \mathcal{L}\left(\theta^{(t+1)}\right) - \mathcal{L}\left(\theta^{(t)}\right) \ge 0
\end{equation}

where we can conclude that the EM algorithm would boost the network performance.

\section{More Qualitative Results}\label{sec:more_quali_res}

We provide more qualitative results to demonstrate the effectiveness of our method; see \cref{fig:qualitative_supp}.

\begin{figure*}[t!]
    \centering
    \begin{subfigure}[t]{0.9\linewidth}
        \centering
        \begin{overpic}
            [width=0.16\linewidth]{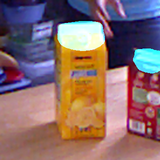}%
            \put(25,105){\small\color{black}{Openable}}
            \put(-20,35){\rotatebox{90}{\color{black}{Ours}}}
        \end{overpic}%
        \begin{overpic}
            [width=0.16\linewidth]{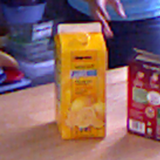}%
            \put(28,105){\small\color{black}{Cuttable}}
        \end{overpic}%
        \begin{overpic}
            [width=0.16\linewidth]{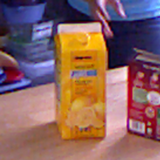}%
            \put(28,105){\small\color{black}{Pourable}}
        \end{overpic}%
        \begin{overpic}
            [width=0.16\linewidth]{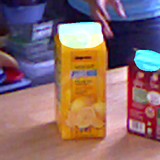}%
            \put(20,105){\small\color{black}{Containable}}
        \end{overpic}%
        \begin{overpic}
            [width=0.16\linewidth]{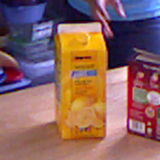}%
            \put(20,105){\small\color{black}{Supportable}}
        \end{overpic}%
        \begin{overpic}
            [width=0.16\linewidth]{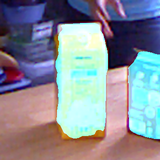}%
            \put(27,105){\small\color{black}{Holdable}}
        \end{overpic}%
        \\%
        \begin{overpic}
            [width=0.16\linewidth]{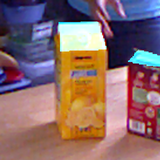}%
            \put(-20,40){\rotatebox{90}{\small\color{black}{GT}}}
        \end{overpic}%
        \includegraphics[width=0.16\linewidth]{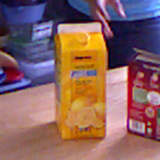}%
        \includegraphics[width=0.16\linewidth]{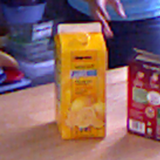}%
        \includegraphics[width=0.16\linewidth]{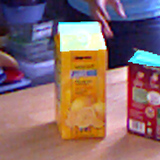}%
        \includegraphics[width=0.16\linewidth]{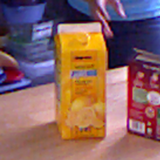}%
        \includegraphics[width=0.16\linewidth]{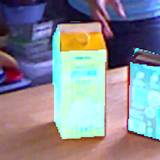}%
    \end{subfigure}%
    \\%
    \begin{subfigure}[t]{0.9\linewidth}
        \centering
        \begin{overpic}
            [width=0.16\linewidth]{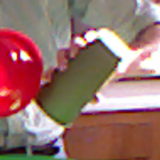}%
            \put(-20,35){\rotatebox{90}{\color{black}{Ours}}}
        \end{overpic}%
        \includegraphics[width=0.16\linewidth]{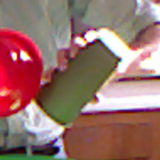}%
        \includegraphics[width=0.16\linewidth]{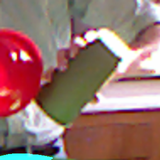}%
        \includegraphics[width=0.16\linewidth]{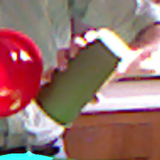}%
        \includegraphics[width=0.16\linewidth]{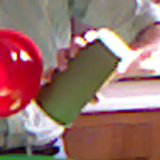}%
        \includegraphics[width=0.16\linewidth]{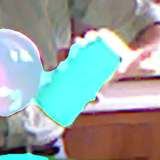}%
        \\%
        \begin{overpic}
            [width=0.16\linewidth]{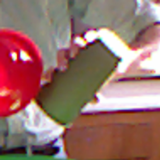}%
            \put(-20,40){\rotatebox{90}{\small\color{black}{GT}}}
        \end{overpic}%
        \includegraphics[width=0.16\linewidth]{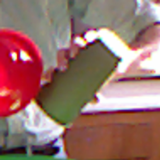}%
        \includegraphics[width=0.16\linewidth]{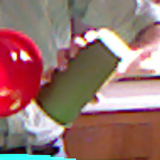}%
        \includegraphics[width=0.16\linewidth]{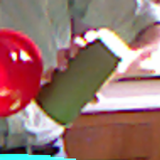}%
        \includegraphics[width=0.16\linewidth]{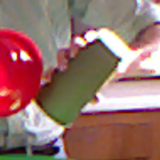}%
        \includegraphics[width=0.16\linewidth]{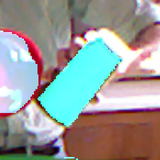}%
    \end{subfigure}%
    \\%
    \begin{subfigure}[t]{0.9\linewidth}
        \centering
        \begin{overpic}
            [width=0.16\linewidth]{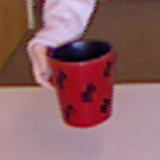}%
            \put(-20,35){\rotatebox{90}{\color{black}{Ours}}}
        \end{overpic}%
        \includegraphics[width=0.16\linewidth]{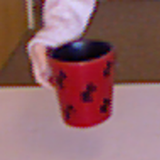}%
        \includegraphics[width=0.16\linewidth]{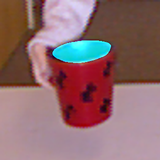}%
        \includegraphics[width=0.16\linewidth]{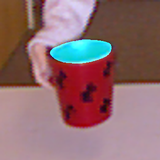}%
        \includegraphics[width=0.16\linewidth]{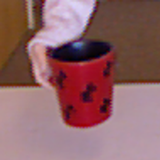}%
        \includegraphics[width=0.16\linewidth]{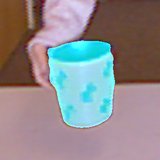}%
        \\%
        \begin{overpic}
            [width=0.16\linewidth]{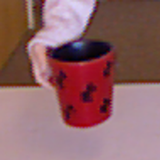}%
            \put(-20,40){\rotatebox{90}{\small\color{black}{GT}}}
        \end{overpic}%
        \includegraphics[width=0.16\linewidth]{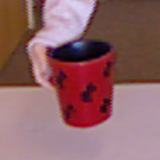}%
        \includegraphics[width=0.16\linewidth]{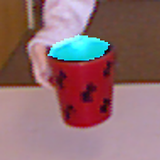}%
        \includegraphics[width=0.16\linewidth]{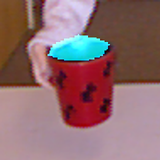}%
        \includegraphics[width=0.16\linewidth]{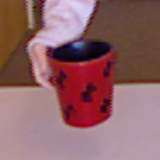}%
        \includegraphics[width=0.16\linewidth]{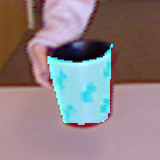}%
    \end{subfigure}%
    \\%
    \begin{subfigure}[t]{0.9\linewidth}
        \centering
        \begin{overpic}
            [width=0.16\linewidth]{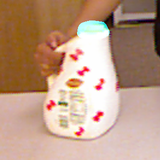}%
            \put(-20,35){\rotatebox{90}{\color{black}{Ours}}}
        \end{overpic}%
        \includegraphics[width=0.16\linewidth]{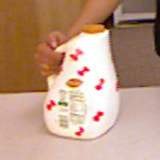}%
        \includegraphics[width=0.16\linewidth]{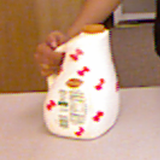}%
        \includegraphics[width=0.16\linewidth]{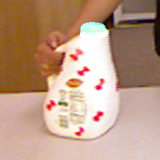}%
        \includegraphics[width=0.16\linewidth]{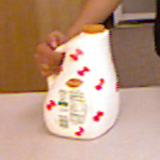}%
        \includegraphics[width=0.16\linewidth]{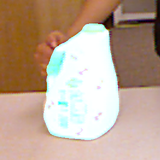}%
        \\%
        \begin{overpic}
            [width=0.16\linewidth]{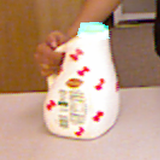}%
            \put(-20,40){\rotatebox{90}{\small\color{black}{GT}}}
        \end{overpic}%
        \includegraphics[width=0.16\linewidth]{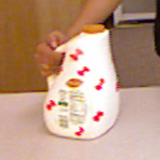}%
        \includegraphics[width=0.16\linewidth]{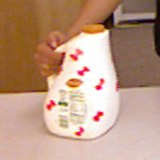}%
        \includegraphics[width=0.16\linewidth]{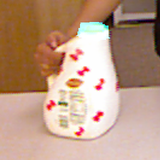}%
        \includegraphics[width=0.16\linewidth]{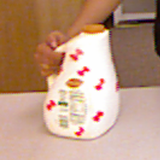}%
        \includegraphics[width=0.16\linewidth]{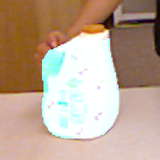}%
    \end{subfigure}%

    \caption{More qualitative results.}
    \label{fig:qualitative_supp}
\end{figure*}

\begin{figure*}[t!]
    \ContinuedFloat
    \centering
    \begin{subfigure}[t]{0.9\linewidth}
        \centering
        \begin{overpic}
            [width=0.16\linewidth]{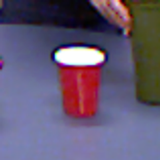}%
            \put(25,105){\small\color{black}{Openable}}
            \put(-20,35){\rotatebox{90}{\color{black}{Ours}}}
        \end{overpic}%
        \begin{overpic}
            [width=0.16\linewidth]{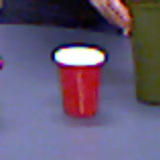}%
            \put(28,105){\small\color{black}{Cuttable}}
        \end{overpic}%
        \begin{overpic}
            [width=0.16\linewidth]{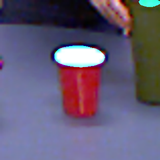}%
            \put(28,105){\small\color{black}{Pourable}}
        \end{overpic}%
        \begin{overpic}
            [width=0.16\linewidth]{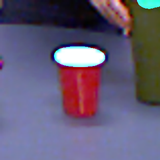}%
            \put(20,105){\small\color{black}{Containable}}
        \end{overpic}%
        \begin{overpic}
            [width=0.16\linewidth]{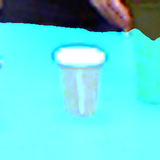}%
            \put(20,105){\small\color{black}{Supportable}}
        \end{overpic}%
        \begin{overpic}
            [width=0.16\linewidth]{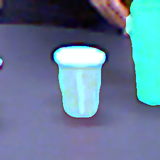}%
            \put(27,105){\small\color{black}{Holdable}}
        \end{overpic}%
        \\%
        \begin{overpic}
            [width=0.16\linewidth]{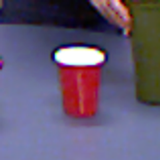}%
            \put(-20,40){\rotatebox{90}{\small\color{black}{GT}}}
        \end{overpic}%
        \includegraphics[width=0.16\linewidth]{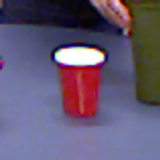}%
        \includegraphics[width=0.16\linewidth]{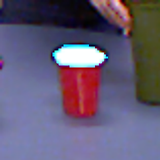}%
        \includegraphics[width=0.16\linewidth]{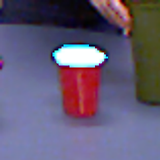}%
        \includegraphics[width=0.16\linewidth]{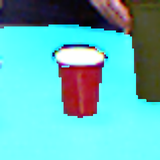}%
        \includegraphics[width=0.16\linewidth]{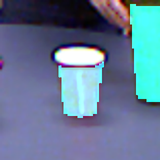}%
    \end{subfigure}%
    \\%
    \begin{subfigure}[t]{0.9\linewidth}
        \centering
        \begin{overpic}
            [width=0.16\linewidth]{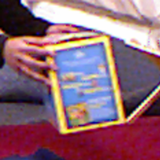}%
            \put(-20,35){\rotatebox{90}{\color{black}{Ours}}}
        \end{overpic}%
        \includegraphics[width=0.16\linewidth]{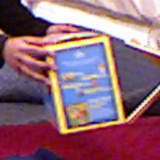}%
        \includegraphics[width=0.16\linewidth]{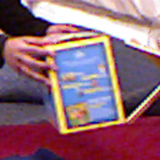}%
        \includegraphics[width=0.16\linewidth]{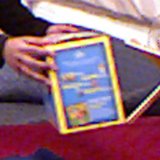}%
        \includegraphics[width=0.16\linewidth]{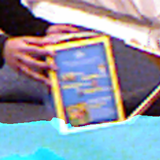}%
        \includegraphics[width=0.16\linewidth]{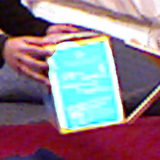}%
        \\%
        \begin{overpic}
            [width=0.16\linewidth]{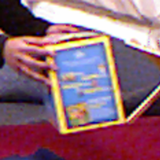}%
            \put(-20,40){\rotatebox{90}{\small\color{black}{GT}}}
        \end{overpic}%
        \includegraphics[width=0.16\linewidth]{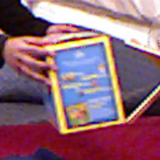}%
        \includegraphics[width=0.16\linewidth]{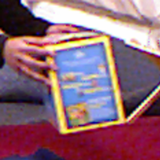}%
        \includegraphics[width=0.16\linewidth]{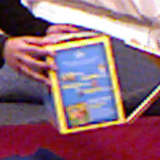}%
        \includegraphics[width=0.16\linewidth]{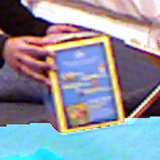}%
        \includegraphics[width=0.16\linewidth]{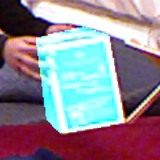}%
    \end{subfigure}%
    \\%
    \begin{subfigure}[t]{0.9\linewidth}
        \centering
        \begin{overpic}
            [width=0.16\linewidth]{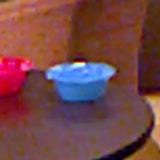}%
            \put(-20,35){\rotatebox{90}{\color{black}{Ours}}}
        \end{overpic}%
        \includegraphics[width=0.16\linewidth]{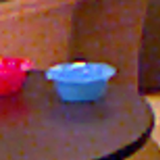}%
        \includegraphics[width=0.16\linewidth]{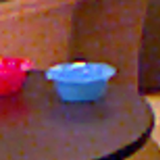}%
        \includegraphics[width=0.16\linewidth]{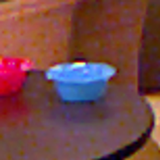}%
        \includegraphics[width=0.16\linewidth]{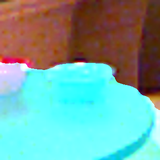}%
        \includegraphics[width=0.16\linewidth]{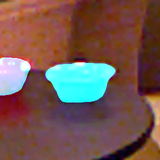}%
        \\%
        \begin{overpic}
            [width=0.16\linewidth]{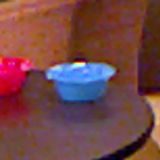}%
            \put(-20,40){\rotatebox{90}{\small\color{black}{GT}}}
        \end{overpic}%
        \includegraphics[width=0.16\linewidth]{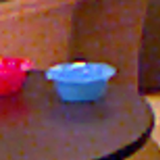}%
        \includegraphics[width=0.16\linewidth]{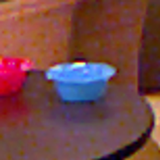}%
        \includegraphics[width=0.16\linewidth]{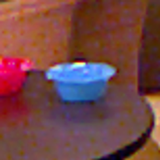}%
        \includegraphics[width=0.16\linewidth]{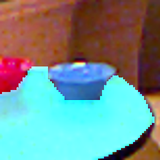}%
        \includegraphics[width=0.16\linewidth]{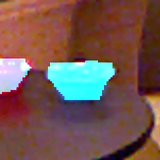}%
    \end{subfigure}%
    \\%
    \begin{subfigure}[t]{0.9\linewidth}
        \centering
        \begin{overpic}
            [width=0.16\linewidth]{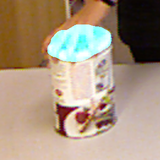}%
            \put(-20,35){\rotatebox{90}{\color{black}{Ours}}}
        \end{overpic}%
        \includegraphics[width=0.16\linewidth]{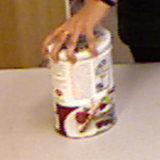}%
        \includegraphics[width=0.16\linewidth]{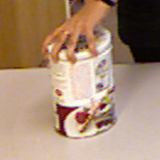}%
        \includegraphics[width=0.16\linewidth]{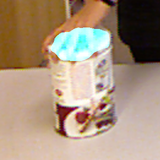}%
        \includegraphics[width=0.16\linewidth]{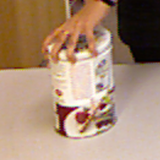}%
        \includegraphics[width=0.16\linewidth]{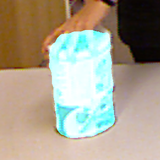}%
        \\%
        \begin{overpic}
            [width=0.16\linewidth]{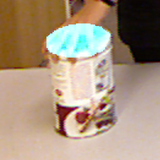}%
            \put(-20,40){\rotatebox{90}{\small\color{black}{GT}}}
        \end{overpic}%
        \includegraphics[width=0.16\linewidth]{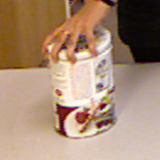}%
        \includegraphics[width=0.16\linewidth]{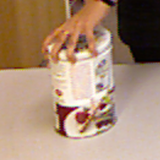}%
        \includegraphics[width=0.16\linewidth]{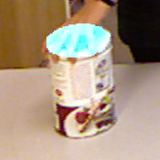}%
        \includegraphics[width=0.16\linewidth]{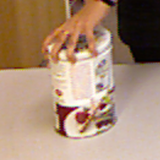}%
        \includegraphics[width=0.16\linewidth]{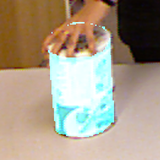}%
    \end{subfigure}%

    \caption{More qualitative results (cont.).}
\end{figure*}

\begin{figure*}[t!]
    \ContinuedFloat
    \centering
    \begin{subfigure}[t]{0.9\linewidth}
        \centering
        \begin{overpic}
            [width=0.16\linewidth]{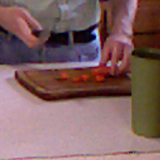}%
            \put(25,105){\small\color{black}{Openable}}
            \put(-20,35){\rotatebox{90}{\color{black}{Ours}}}
        \end{overpic}%
        \begin{overpic}
            [width=0.16\linewidth]{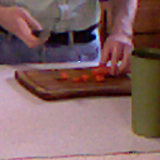}%
            \put(28,105){\small\color{black}{Cuttable}}
        \end{overpic}%
        \begin{overpic}
            [width=0.16\linewidth]{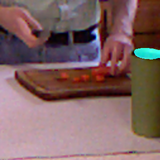}%
            \put(28,105){\small\color{black}{Pourable}}
        \end{overpic}%
        \begin{overpic}
            [width=0.16\linewidth]{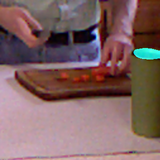}%
            \put(20,105){\small\color{black}{Containable}}
        \end{overpic}%
        \begin{overpic}
            [width=0.16\linewidth]{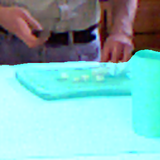}%
            \put(20,105){\small\color{black}{Supportable}}
        \end{overpic}%
        \begin{overpic}
            [width=0.16\linewidth]{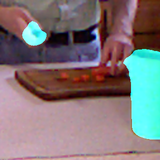}%
            \put(27,105){\small\color{black}{Holdable}}
        \end{overpic}%
        \\%
        \begin{overpic}
            [width=0.16\linewidth]{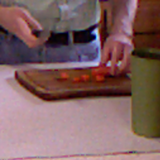}%
            \put(-20,40){\rotatebox{90}{\small\color{black}{GT}}}
        \end{overpic}%
        \includegraphics[width=0.16\linewidth]{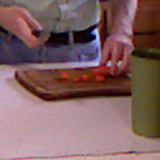}%
        \includegraphics[width=0.16\linewidth]{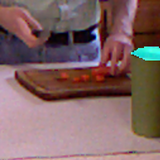}%
        \includegraphics[width=0.16\linewidth]{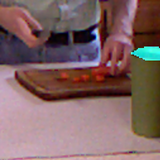}%
        \includegraphics[width=0.16\linewidth]{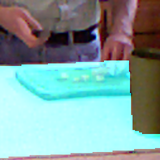}%
        \includegraphics[width=0.16\linewidth]{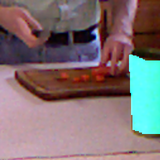}%
    \end{subfigure}%
    \\%
    \begin{subfigure}[t]{0.9\linewidth}
        \centering
        \begin{overpic}
            [width=0.16\linewidth]{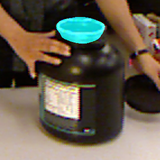}%
            \put(-20,35){\rotatebox{90}{\color{black}{Ours}}}
        \end{overpic}%
        \includegraphics[width=0.16\linewidth]{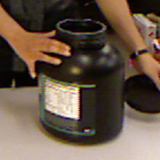}%
        \includegraphics[width=0.16\linewidth]{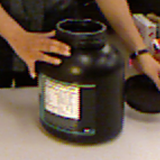}%
        \includegraphics[width=0.16\linewidth]{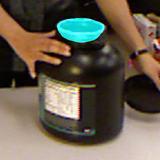}%
        \includegraphics[width=0.16\linewidth]{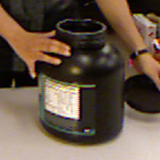}%
        \includegraphics[width=0.16\linewidth]{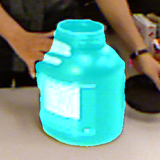}%
        \\%
        \begin{overpic}
            [width=0.16\linewidth]{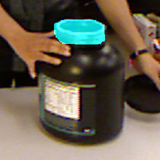}%
            \put(-20,40){\rotatebox{90}{\small\color{black}{GT}}}
        \end{overpic}%
        \includegraphics[width=0.16\linewidth]{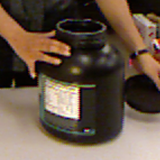}%
        \includegraphics[width=0.16\linewidth]{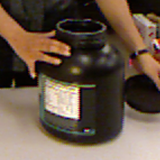}%
        \includegraphics[width=0.16\linewidth]{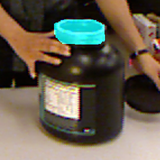}%
        \includegraphics[width=0.16\linewidth]{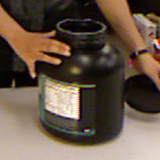}%
        \includegraphics[width=0.16\linewidth]{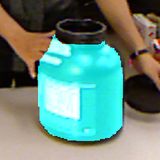}%
    \end{subfigure}%
    \\%
    \begin{subfigure}[t]{0.9\linewidth}
        \centering
        \begin{overpic}
            [width=0.16\linewidth]{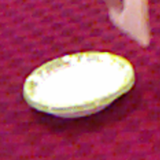}%
            \put(-20,35){\rotatebox{90}{\color{black}{Ours}}}
        \end{overpic}%
        \includegraphics[width=0.16\linewidth]{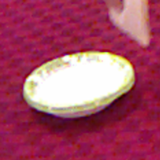}%
        \includegraphics[width=0.16\linewidth]{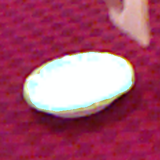}%
        \includegraphics[width=0.16\linewidth]{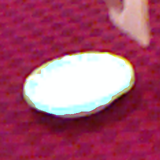}%
        \includegraphics[width=0.16\linewidth]{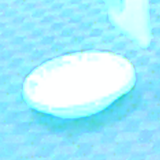}%
        \includegraphics[width=0.16\linewidth]{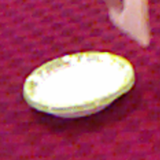}%
        \\%
        \begin{overpic}
            [width=0.16\linewidth]{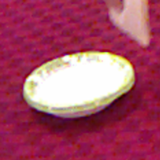}%
            \put(-20,40){\rotatebox{90}{\small\color{black}{GT}}}
        \end{overpic}%
        \includegraphics[width=0.16\linewidth]{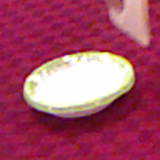}%
        \includegraphics[width=0.16\linewidth]{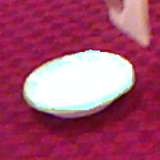}%
        \includegraphics[width=0.16\linewidth]{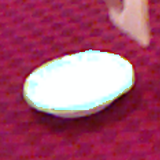}%
        \includegraphics[width=0.16\linewidth]{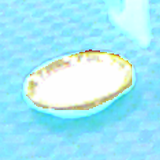}%
        \includegraphics[width=0.16\linewidth]{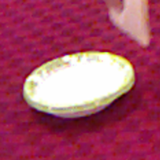}%
    \end{subfigure}%
    \\%
    \begin{subfigure}[t]{0.9\linewidth}
        \centering
        \begin{overpic}
            [width=0.16\linewidth]{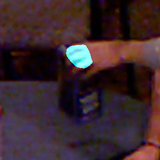}%
            \put(-20,35){\rotatebox{90}{\color{black}{Ours}}}
        \end{overpic}%
        \includegraphics[width=0.16\linewidth]{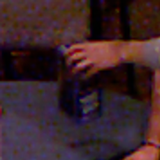}%
        \includegraphics[width=0.16\linewidth]{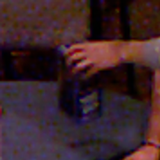}%
        \includegraphics[width=0.16\linewidth]{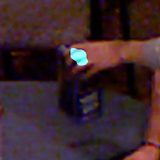}%
        \includegraphics[width=0.16\linewidth]{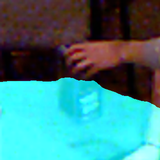}%
        \includegraphics[width=0.16\linewidth]{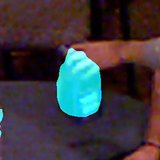}%
        \\%
        \begin{overpic}
            [width=0.16\linewidth]{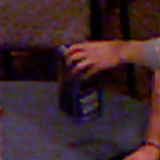}%
            \put(-20,40){\rotatebox{90}{\small\color{black}{GT}}}
        \end{overpic}%
        \includegraphics[width=0.16\linewidth]{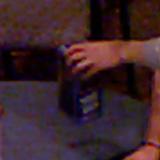}%
        \includegraphics[width=0.16\linewidth]{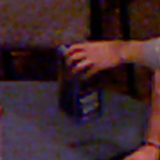}%
        \includegraphics[width=0.16\linewidth]{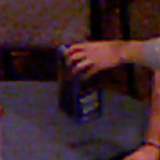}%
        \includegraphics[width=0.16\linewidth]{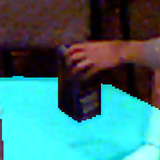}%
        \includegraphics[width=0.16\linewidth]{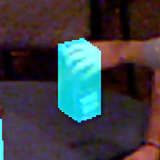}%
    \end{subfigure}%

    \caption{More qualitative results (cont.).}
\end{figure*}

\begin{figure*}[t!]
    \ContinuedFloat
    \centering
    \begin{subfigure}[t]{0.9\linewidth}
        \centering
        \begin{overpic}
            [width=0.16\linewidth]{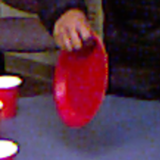}%
            \put(25,105){\small\color{black}{Openable}}
            \put(-20,35){\rotatebox{90}{\color{black}{Ours}}}
        \end{overpic}%
        \begin{overpic}
            [width=0.16\linewidth]{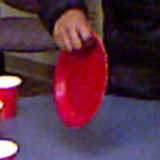}%
            \put(28,105){\small\color{black}{Cuttable}}
        \end{overpic}%
        \begin{overpic}
            [width=0.16\linewidth]{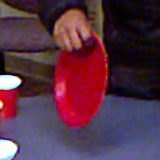}%
            \put(28,105){\small\color{black}{Pourable}}
        \end{overpic}%
        \begin{overpic}
            [width=0.16\linewidth]{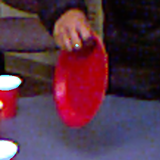}%
            \put(20,105){\small\color{black}{Containable}}
        \end{overpic}%
        \begin{overpic}
            [width=0.16\linewidth]{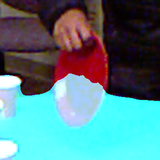}%
            \put(20,105){\small\color{black}{Supportable}}
        \end{overpic}%
        \begin{overpic}
            [width=0.16\linewidth]{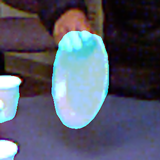}%
            \put(27,105){\small\color{black}{Holdable}}
        \end{overpic}%
        \\%
        \begin{overpic}
            [width=0.16\linewidth]{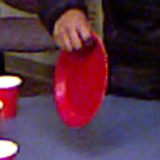}%
            \put(-20,40){\rotatebox{90}{\small\color{black}{GT}}}
        \end{overpic}%
        \includegraphics[width=0.16\linewidth]{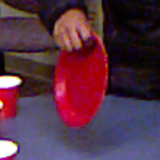}%
        \includegraphics[width=0.16\linewidth]{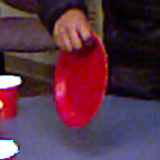}%
        \includegraphics[width=0.16\linewidth]{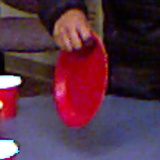}%
        \includegraphics[width=0.16\linewidth]{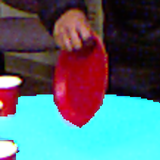}%
        \includegraphics[width=0.16\linewidth]{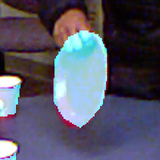}%
    \end{subfigure}%
    \\%
    \begin{subfigure}[t]{0.9\linewidth}
        \centering
        \begin{overpic}
            [width=0.16\linewidth]{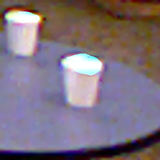}%
            \put(-20,35){\rotatebox{90}{\color{black}{Ours}}}
        \end{overpic}%
        \includegraphics[width=0.16\linewidth]{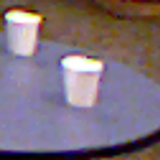}%
        \includegraphics[width=0.16\linewidth]{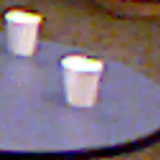}%
        \includegraphics[width=0.16\linewidth]{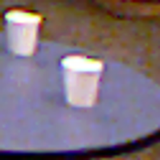}%
        \includegraphics[width=0.16\linewidth]{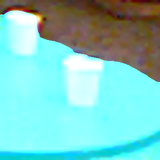}%
        \includegraphics[width=0.16\linewidth]{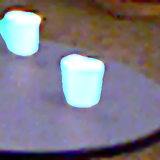}%
        \\%
        \begin{overpic}
            [width=0.16\linewidth]{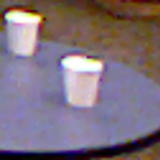}%
            \put(-20,40){\rotatebox{90}{\small\color{black}{GT}}}
        \end{overpic}%
        \includegraphics[width=0.16\linewidth]{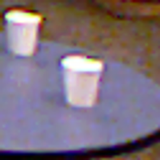}%
        \includegraphics[width=0.16\linewidth]{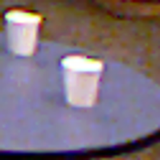}%
        \includegraphics[width=0.16\linewidth]{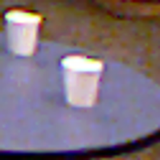}%
        \includegraphics[width=0.16\linewidth]{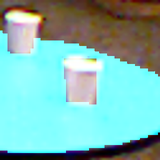}%
        \includegraphics[width=0.16\linewidth]{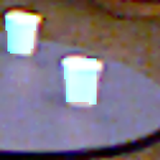}%
    \end{subfigure}%
    \\%
    \begin{subfigure}[t]{0.9\linewidth}
        \centering
        \begin{overpic}
            [width=0.16\linewidth]{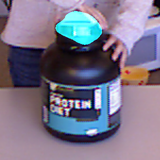}%
            \put(-20,35){\rotatebox{90}{\color{black}{Ours}}}
        \end{overpic}%
        \includegraphics[width=0.16\linewidth]{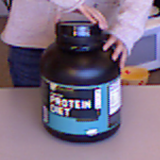}%
        \includegraphics[width=0.16\linewidth]{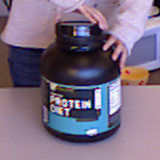}%
        \includegraphics[width=0.16\linewidth]{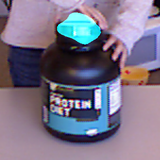}%
        \includegraphics[width=0.16\linewidth]{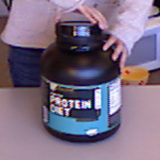}%
        \includegraphics[width=0.16\linewidth]{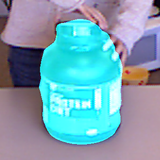}%
        \\%
        \begin{overpic}
            [width=0.16\linewidth]{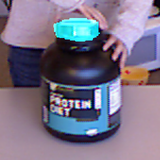}%
            \put(-20,40){\rotatebox{90}{\small\color{black}{GT}}}
        \end{overpic}%
        \includegraphics[width=0.16\linewidth]{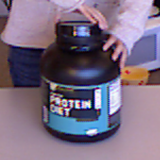}%
        \includegraphics[width=0.16\linewidth]{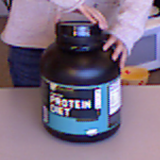}%
        \includegraphics[width=0.16\linewidth]{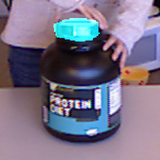}%
        \includegraphics[width=0.16\linewidth]{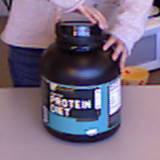}%
        \includegraphics[width=0.16\linewidth]{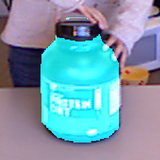}%
    \end{subfigure}%
    \\%
    \begin{subfigure}[t]{0.9\linewidth}
        \centering
        \begin{overpic}
            [width=0.16\linewidth]{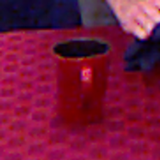}%
            \put(-20,35){\rotatebox{90}{\color{black}{Ours}}}
        \end{overpic}%
        \includegraphics[width=0.16\linewidth]{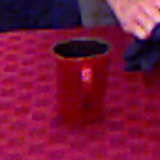}%
        \includegraphics[width=0.16\linewidth]{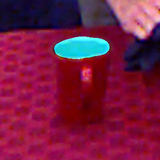}%
        \includegraphics[width=0.16\linewidth]{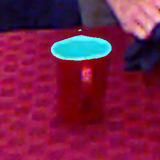}%
        \includegraphics[width=0.16\linewidth]{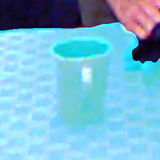}%
        \includegraphics[width=0.16\linewidth]{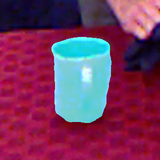}%
        \\%
        \begin{overpic}
            [width=0.16\linewidth]{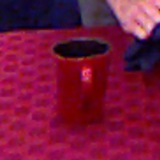}%
            \put(-20,40){\rotatebox{90}{\small\color{black}{GT}}}
        \end{overpic}%
        \includegraphics[width=0.16\linewidth]{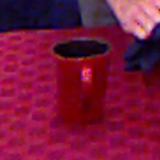}%
        \includegraphics[width=0.16\linewidth]{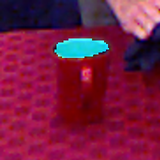}%
        \includegraphics[width=0.16\linewidth]{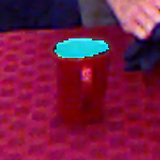}%
        \includegraphics[width=0.16\linewidth]{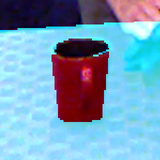}%
        \includegraphics[width=0.16\linewidth]{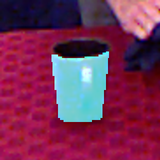}%
    \end{subfigure}%

    \caption{More qualitative results (cont.).}
\end{figure*}

\begin{figure*}[t!]
    \ContinuedFloat
    \centering
    \begin{subfigure}[t]{0.9\linewidth}
        \centering
        \begin{overpic}
            [width=0.16\linewidth]{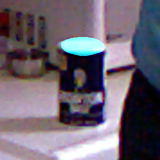}%
            \put(25,105){\small\color{black}{Openable}}
            \put(-20,35){\rotatebox{90}{\color{black}{Ours}}}
        \end{overpic}%
        \begin{overpic}
            [width=0.16\linewidth]{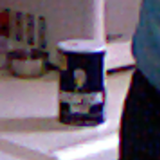}%
            \put(28,105){\small\color{black}{Cuttable}}
        \end{overpic}%
        \begin{overpic}
            [width=0.16\linewidth]{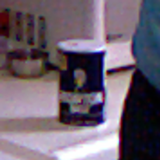}%
            \put(28,105){\small\color{black}{Pourable}}
        \end{overpic}%
        \begin{overpic}
            [width=0.16\linewidth]{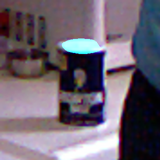}%
            \put(20,105){\small\color{black}{Containable}}
        \end{overpic}%
        \begin{overpic}
            [width=0.16\linewidth]{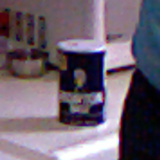}%
            \put(20,105){\small\color{black}{Supportable}}
        \end{overpic}%
        \begin{overpic}
            [width=0.16\linewidth]{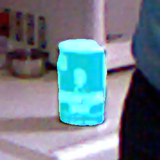}%
            \put(27,105){\small\color{black}{Holdable}}
        \end{overpic}%
        \\%
        \begin{overpic}
            [width=0.16\linewidth]{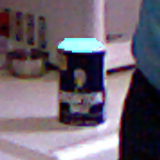}%
            \put(-20,40){\rotatebox{90}{\small\color{black}{GT}}}
        \end{overpic}%
        \includegraphics[width=0.16\linewidth]{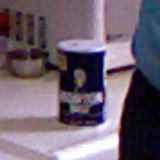}%
        \includegraphics[width=0.16\linewidth]{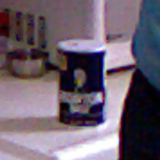}%
        \includegraphics[width=0.16\linewidth]{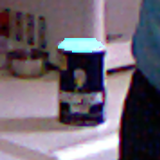}%
        \includegraphics[width=0.16\linewidth]{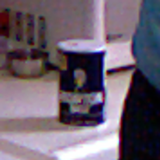}%
        \includegraphics[width=0.16\linewidth]{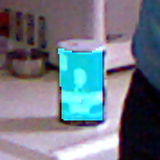}%
    \end{subfigure}%
    \\%
    \begin{subfigure}[t]{0.9\linewidth}
        \centering
        \begin{overpic}
            [width=0.16\linewidth]{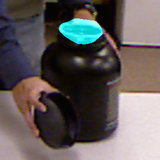}%
            \put(-20,35){\rotatebox{90}{\color{black}{Ours}}}
        \end{overpic}%
        \includegraphics[width=0.16\linewidth]{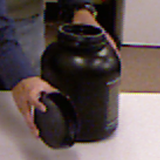}%
        \includegraphics[width=0.16\linewidth]{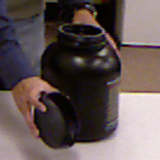}%
        \includegraphics[width=0.16\linewidth]{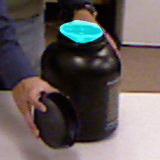}%
        \includegraphics[width=0.16\linewidth]{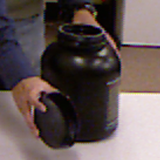}%
        \includegraphics[width=0.16\linewidth]{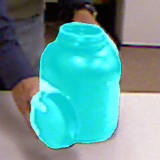}%
        \\%
        \begin{overpic}
            [width=0.16\linewidth]{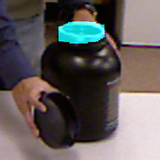}%
            \put(-20,40){\rotatebox{90}{\small\color{black}{GT}}}
        \end{overpic}%
        \includegraphics[width=0.16\linewidth]{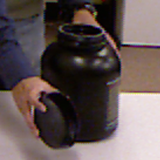}%
        \includegraphics[width=0.16\linewidth]{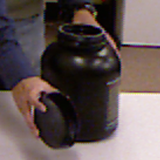}%
        \includegraphics[width=0.16\linewidth]{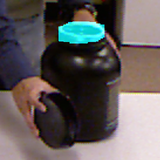}%
        \includegraphics[width=0.16\linewidth]{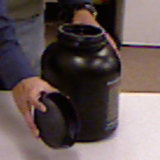}%
        \includegraphics[width=0.16\linewidth]{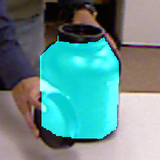}%
    \end{subfigure}%
    \\%
    \begin{subfigure}[t]{0.9\linewidth}
        \centering
        \begin{overpic}
            [width=0.16\linewidth]{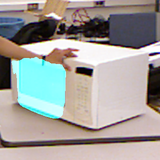}%
            \put(-20,35){\rotatebox{90}{\color{black}{Ours}}}
        \end{overpic}%
        \includegraphics[width=0.16\linewidth]{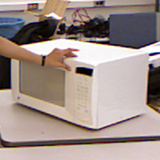}%
        \includegraphics[width=0.16\linewidth]{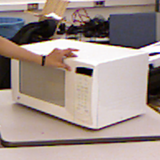}%
        \includegraphics[width=0.16\linewidth]{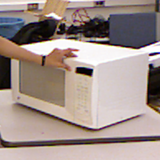}%
        \includegraphics[width=0.16\linewidth]{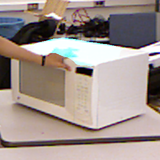}%
        \includegraphics[width=0.16\linewidth]{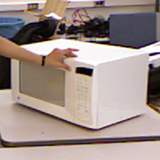}%
        \\%
        \begin{overpic}
            [width=0.16\linewidth]{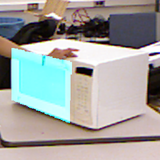}%
            \put(-20,40){\rotatebox{90}{\small\color{black}{GT}}}
        \end{overpic}%
        \includegraphics[width=0.16\linewidth]{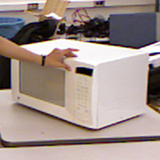}%
        \includegraphics[width=0.16\linewidth]{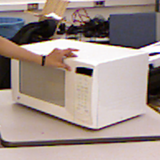}%
        \includegraphics[width=0.16\linewidth]{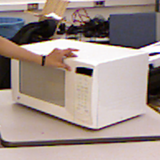}%
        \includegraphics[width=0.16\linewidth]{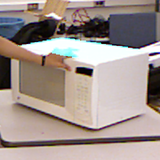}%
        \includegraphics[width=0.16\linewidth]{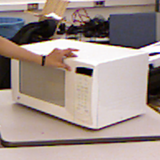}%
    \end{subfigure}%
    \\%
    \begin{subfigure}[t]{0.9\linewidth}
        \centering
        \begin{overpic}
            [width=0.16\linewidth]{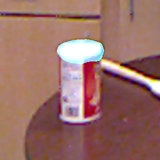}%
            \put(-20,35){\rotatebox{90}{\color{black}{Ours}}}
        \end{overpic}%
        \includegraphics[width=0.16\linewidth]{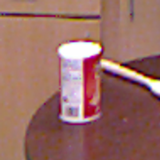}%
        \includegraphics[width=0.16\linewidth]{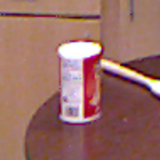}%
        \includegraphics[width=0.16\linewidth]{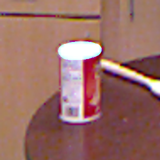}%
        \includegraphics[width=0.16\linewidth]{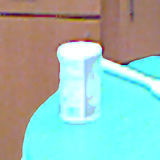}%
        \includegraphics[width=0.16\linewidth]{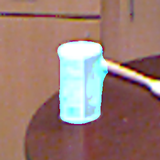}%
        \\%
        \begin{overpic}
            [width=0.16\linewidth]{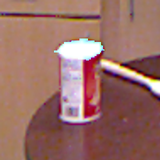}%
            \put(-20,40){\rotatebox{90}{\small\color{black}{GT}}}
        \end{overpic}%
        \includegraphics[width=0.16\linewidth]{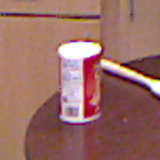}%
        \includegraphics[width=0.16\linewidth]{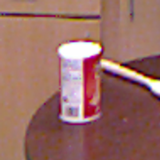}%
        \includegraphics[width=0.16\linewidth]{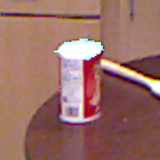}%
        \includegraphics[width=0.16\linewidth]{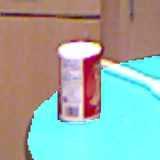}%
        \includegraphics[width=0.16\linewidth]{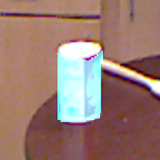}%
    \end{subfigure}%

    \caption{More qualitative results (cont.).}
\end{figure*}

\end{document}